\documentclass{ejm}

\usepackage{hyperref}

\usepackage[usenames,dvipsnames]{xcolor}
\usepackage{graphicx}
\usepackage[center]{subfigure}
\usepackage{amsmath}
\usepackage{cleveref}
\usepackage{blindtext}
\usepackage{array}
\usepackage{spverbatim}
\usepackage{listings}
\usepackage{fancyvrb}
\usepackage{everysel}
\usepackage{cite}

\newcommand{\p}[1]{\boldsymbol{#1}}

\newcommand{\nbauthors}{651}
\newcommand{\nbdocuments}{14947}
\newcommand{\justify}{
\fontdimen2\font=0.4em
\fontdimen3\font=0.2em
\fontdimen4\font=0.1em
\fontdimen7\font=0.1em
\hyphenchar\font=`\-
}
\usepackage{caption}
\usepackage{fontenc}
\newcommand{\todefine}[1]{``{#1}''}
\newcommand{\quoting}[1]{``{#1}''}
\newcommand{\emdash}{{\normalfont ---}}

\usepackage[section]{placeins}

\let\realverbatim=\verbatim
\let\realendverbatim=\endverbatim
\renewcommand\verbatim{\par\addvspace{6pt plus 2pt minus 1pt}\realverbatim}
\renewcommand\endverbatim{\realendverbatim\addvspace{6pt plus 2pt minus 1pt}}

\ifprodtf \else
  \checkfont{eurm10}
  \iffontfound
    \IfFileExists{upmath.sty}
      {\typeout{^^JFound AMS Euler Roman fonts on the system,
                   using the 'upmath' package.^^J}%
       \usepackage{upmath}}
      {\typeout{^^JFound AMS Euler Roman fonts on the system, but you
                   don't seem to have the}%
       \typeout{'upmath' package installed. EJM.cls can take advantage
                 of these fonts,^^Jif you use 'upmath' package.^^J}%
       \providecommand\mathup[1]{##1}%
       \providecommand\mathbup[1]{##1}%
      }
  \else
    \providecommand\mathup[1]{#1}%
    \providecommand\mathbup[1]{#1}%
  \fi
\fi

\ifprodtf \else
  \checkfont{msam10}
  \iffontfound
    \IfFileExists{amssymb.sty}
      {\typeout{^^JFound AMS Symbol fonts on the system, using the
                'amssymb' package.^^J}%
       \usepackage{amssymb}%
         \let\leq=\leqslant
         \let\geq=\geqslant
      }{}
  \fi
\fi

\ifprodtf \else
  \IfFileExists{amsbsy.sty}
    {\typeout{^^JFound the 'amsbsy' package on the system, using it.^^J}%
     \usepackage{amsbsy}}
    {\providecommand\boldsymbol[1]{\mbox{\boldmath $##1$}}}
\fi

\newsavebox{\astrutbox}
\sbox{\astrutbox}{\rule[-5pt]{0pt}{20pt}}

\graphicspath{{figs/}}

\newdefinition{definition}[theorem]{Definition}

\title[Pull out all the stops]{Pull out all the stops: \\ Textual analysis via punctuation sequences}

\author[A. N. M. Darmon et al.]{%
  ALEXANDRA\ns N.\ns M.\ns D\ls A\ls R\ls M\ls O\ls N$\,^1$,\ns
  MARYA\ns B\ls A\ls Z\ls Z\ls I$\,^{1, 2, 3}$,\ns
  SAM\ns D.\ns H\ls O\ls W\ls I\ls S\ls O\ls N$\,^{1},$
\and
  MASON\ns A.\ns P\ls O\ls R\ls T\ls E\ls R$\,^{1, 4}$\ns
}
\affiliation{
$^1\,$Oxford Centre for Industrial and Applied Mathematics,
Mathematical Institute, University of Oxford, Oxford OX2 6GG, United Kingdom\\
 $^2\,$The Alan Turing Institute, London NW1 2DB, United Kingdom\\
 $^3\,$Warwick Mathematics Institute, University of Warwick,
 Coventry CV4 7AL, United Kingdom\\
 $^4\,$Department of Mathematics,
 University of California, Los Angeles, Los Angeles, California 90095, USA
 }

\pagerange{\pageref{firstpage}--\pageref{lastpage}}

\begin{document}

\label{firstpage}
\maketitle
\texttt{\justify "I'm tired of wasting letters when punctuation will do, period." \\ 
\indent \indent \indent \indent \indent \indent \indent \indent \indent \indent \indent \indent
{\normalfont \emdash{}\;Steve Martin, \textit{Twitter}, 2011}}
\vspace{2em}

\begin{abstract}

Whether enjoying the lucid prose of a favorite author or slogging through some other writer's cumbersome, heavy-set prattle (full of parentheses, em dashes, compound adjectives, and Oxford commas), readers will notice stylistic signatures not only in word choice and grammar, but also in punctuation itself. Indeed, visual sequences of punctuation from different authors produce marvelously different (and visually striking) sequences. Punctuation is a largely overlooked stylistic feature in ``stylometry'', the quantitative analysis of written text. In this paper, we examine punctuation sequences in a corpus of literary documents and ask the following questions: Are the properties of such sequences a distinctive feature of different authors? Is it possible to distinguish literary genres based on their punctuation sequences? Do the punctuation styles of authors evolve over time? Are we on to something interesting in trying to do stylometry without words, or are we full of sound and fury (signifying nothing)?

\end{abstract}




\section{Introduction}
\label{sec:intro}

\begin{figure}[t]
\centering
\subfigure[Punctuation sequence: \textit{Sharing Her Crime}, Agnes May Fleming]{
\includegraphics[width = 3cm]{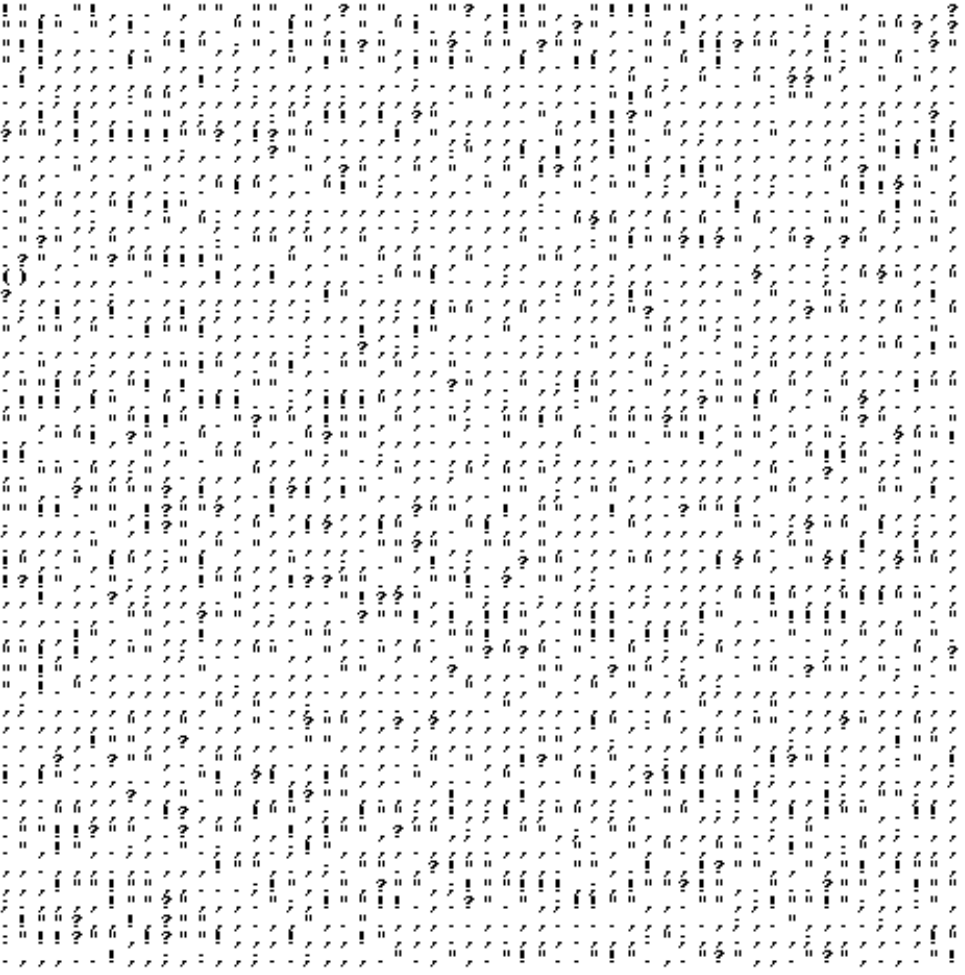}}
\subfigure[Heat map: \textit{Sharing Her Crime}, Agnes May Fleming]{
\includegraphics[width =   3cm]{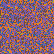}}
\hspace{4mm}
\subfigure[Punctuation sequence: \textit{King Lear}, William Shakespeare]{
\includegraphics[width = 3cm]{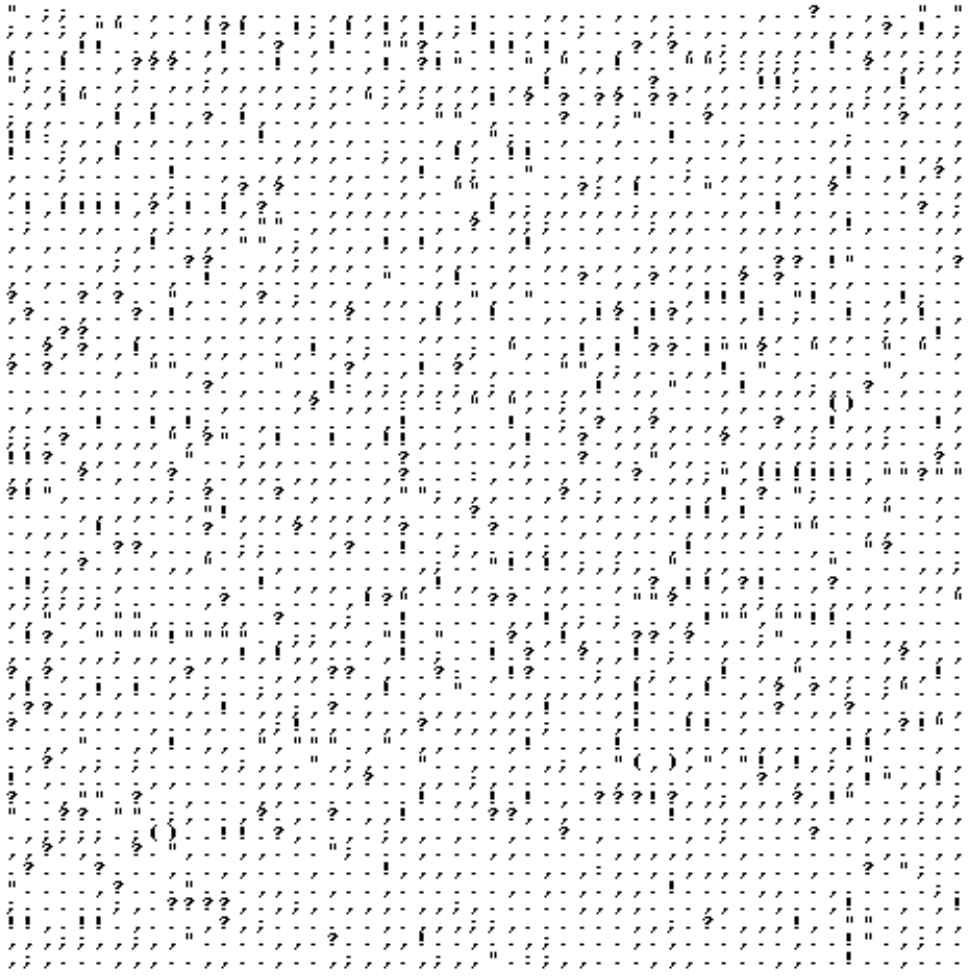}}
\subfigure[Heat map: \textit{King Lear}, William Shakespeare]{
\includegraphics[width = 3cm]{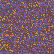}}

\centering\includegraphics[width = 4cm]{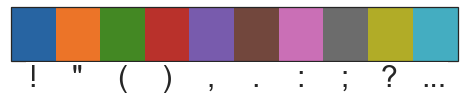}

\subfigure[Punctuation sequence: \textit{The History of Mr. Polly}, Herbert George Wells]{
\includegraphics[width = 3cm]{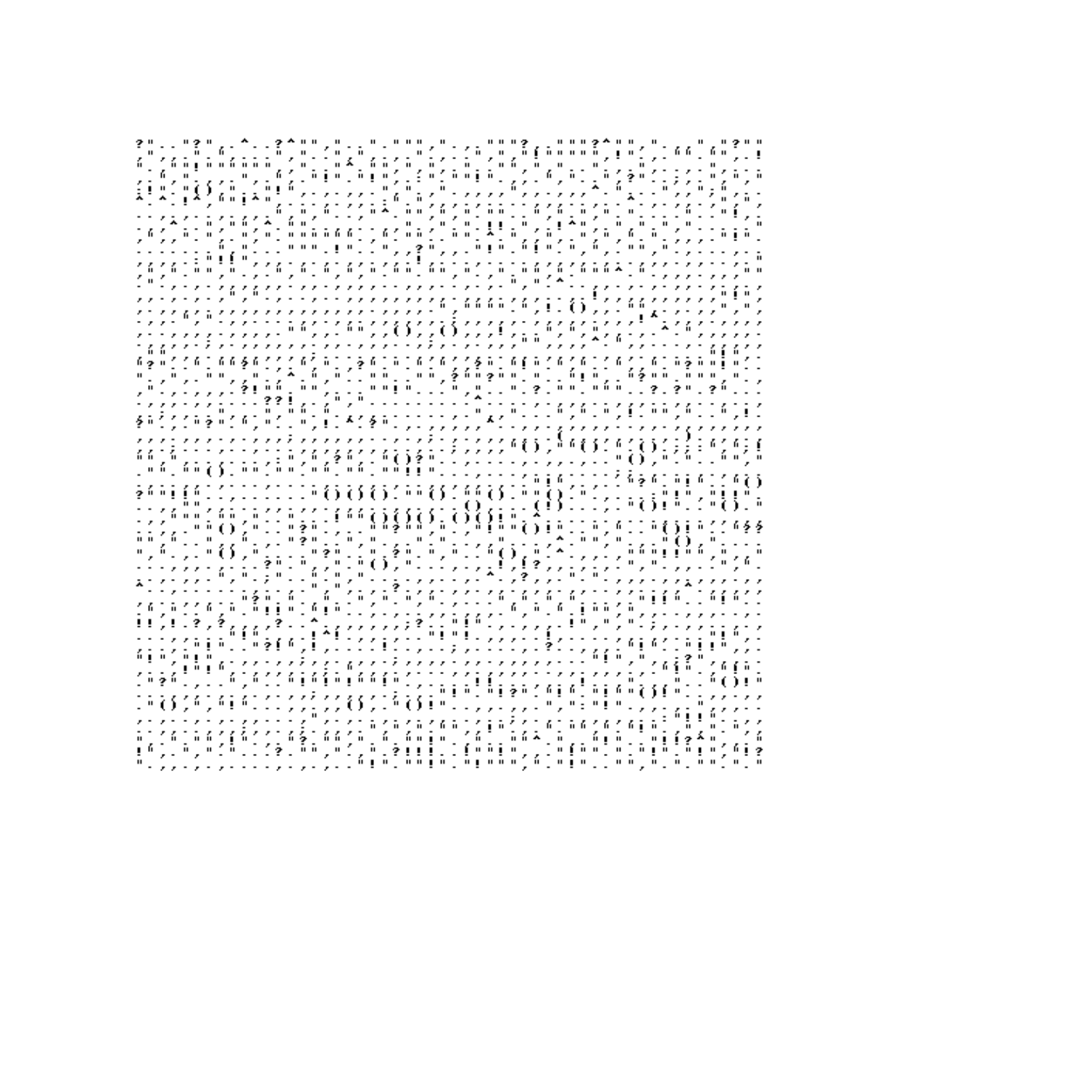}}
\subfigure[Heat map: \textit{The History of Mr. Polly}, Herbert George Wells]{
\includegraphics[width = 3cm]{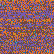}}
\caption[]{(a,c,e) Excerpts of ordered punctuation sequences and (b,d,f) corresponding heat maps for books by three different authors: (a,b) Agnes May Fleming; (c,d) William Shakespeare; and (e,f) Herbert George Wells. Each depicted punctuation sequence consists of $3000$ successive punctuation marks starting from the midpoint of the full punctuation sequence of the corresponding document. The color bar gives the mapping between punctuation marks and colors.
 }
\label{fig:raw_sequence}
\end{figure}


\noindent \texttt{"Yesterday Mr. Hall wrote that the printer's proof-reader was improving my 
punctuation for me, \& I telegraphed orders to have him shot without giving 
him time to pray."\newline
\indent \indent \indent \indent \indent \indent \indent \indent \indent \indent \indent \indent	{\normalfont\emdash{}\;Mark Twain, \emph{Letter to W. Howells}, 1889
}}
\vspace{2em}


\begin{center}
\bf{ ( , , ) . ; . , " " , : ( , ) ; , ? ; , ? ?}
\end{center}

\medskip
\medskip

The sequence of punctuation marks above is what remains of this opening paragraph of our paper (but, to avoid recursion, without the sequence itself) after we remove all of the words. It is perhaps hard to credit that such a minimal sequence encodes any useful information at all; yet it does. In this paper, we investigate the information content of ``de-worded'' documents, asking questions like the following: Do authors have identifiable punctuation styles (see~\cref{fig:raw_sequence}, which was inspired by the visualizations from~\cite{Calhoun2016blog,Calhoun2016code}); if so, can we use them to attribute texts to authors? Do different genres of text differ in their punctuation styles; if so, how? How has punctuation usage evolved over the last few centuries? 

In the present paper, we study sequences of punctuation marks (see~\cref{fig:raw_sequence}) and the number of words that separate punctuation marks. We use Project Gutenberg~\cite{Gutenberg1971} to obtain a large literary corpus. We do not attempt to distinguish between an editor's style and an author's style for the documents in our corpus; doing so for a large corpus in an automated way is a daunting challenge, and we leave it for future efforts. In our work, we investigate whether it is possible to algorithmically assign documents to their authors, irrespective of the documents' edition(s). For ease of writing, we associate documents to authors rather than to both authors and editors throughout our paper, although we recognize that a document's writing and punctuation style can be (and usually is) a product of both.   

Our paper contributes to research areas such as computational linguistics and stylometry. \textit{Computational linguistics} is a research area that, broadly speaking, focuses on the development of computational approaches for processing and analyzing natural language. \textit{Stylometry}, a part of computational linguistics --- as well as cultural analytics, in the broader context of digital humanities --- encompasses quantitative analysis of written text, with the goal of characterizing authorship or other characteristics~\cite{Stamatatos2009, Neal2017, Holmes1997}. Some of the earliest attempts at quantifying the writing style of a document include Mendenhall's work on William Shakespeare's plays in
1887~\cite{Mendenhall1887} and Mosteller et al.'s work on \textit{The Federalist Papers} in 1964~\cite{Mosteller1964}. The latter is often
regarded as the foundation of computer-assisted stylometry (in contrast with methods based on human expertise)~\cite{Stamatatos2009, Neal2017}. Uses of stylometry include (1) authorship attribution, recognition, or detection (which aims to determine whether a document was written by a given author); (2) authorship verification (which aims to determine whether a set of documents were written by the same author); (3) plagiarism detection (which aims to determine similarities between two documents); (4) authorship profiling (which aims to determine certain demographics, such as gender, or other characteristics without directly identifying an author);\footnote{For an example of ``quantitative profiling'', see Neidorf et al.~\cite{Neidorf2019}, who used stylometry to investigate stylistic features (some of which are punctuation-like, as discussed in \url{https://arstechnica.com/science/2019/04/tolkien-was-right-scholars-conclude-beowulf-likely-the-work-of-single-author/}) of {\it Beowulf} and concluded that it is likely the work of a single author.} (5) stylochronometry (which is the study and detection of changes in authorial style over time); and (6) adversarial stylometry (which aims to evade authorship attribution via alteration of style). 

There has been extensive work on author recognition using a wide variety of stylometric features, including ``lexical features'' (e.g., number of words and
mean sentence length), ``syntactic features'' (e.g., frequency of different punctuation marks), ``semantic features'' (e.g., synonyms), and ``structural features'' (e.g., paragraph length and number of words per paragraph). Two common stylometric features for author recognition are ``$n$-grams'' (e.g., in the form of $n$ contiguous words or characters) and ``function words'' (e.g., pronouns, prepositions, and auxiliary verbs). In this paper, in contrast to prior work, we focus on punctuation, rather than on words or letters. We explore several stylometric tasks through the lens of punctuation, illustrating their distinctive role in text.

According to the definition in~\cite{Lawler2006}, \textit{punctuation} refers to the various systems of dots and other marks that accompany letters as part of a writing system. Punctuation is distinct from \textit{diacritic marks}, which are typically modifications of individual letters (e.g., \c{c}, \"{o}, and \H{o}) and \textit{logographs}, which are symbolic representations of lexical items (e.g., \# and \&). Other common symbols, such as the slash to indicate alternation (e.g., and/or) and the asterisk ``\;*\;'', do not fall squarely into one of these categories, but they are not considered to be true punctuation marks~\cite{Lawler2006}. Common punctuation marks are the period (i.e., full stop) ``\;.\;''; the comma ``\;,\;''; the colon ``\;:\;''; the semicolon ``\;;\;''; the left and right parentheses, ``\;(\;'' and ``\;)\;'';  the question mark ``\;?\;''; the exclamation point (which is also called the exclamation mark) ``\;!\;''; the hyphen ``\;-\;''; the en dash ``\;--\;''; the em dash ``\;---\;''; the opening and closing single quotation marks (i.e., inverted commas), ``\;\;`\;\;'' and ``\;\;'\;\;''; the opening and closing double quotation marks (which are also known as inverted commas), ``\;\;``\;\;'', and ``\;\;''\;\;''; the apostrophe ``\;\;'\;\;''; and the ellipsis ``\;...\;''. 

The aforementioned punctuation set (with minor variations) is used today in a large number of alphabetic writing systems and alphabetic languages~\cite{Lawler2006}. In this sense, for a large number of languages, punctuation is a ``supra-linguistic'' representational system. However, punctuation varies significantly across individuals, and there is no consensus on how it should be used~\cite{Fowler1906, Pullum2001, Lewis1979, Truss2004, Nunberg1990}; authors, editors, and typesetters can sometimes get into emphatic disagreements about it.\footnote{Not that any of us would ever descend to this.} Accordingly, as a representational system, punctuation is not standardized, and it may never achieve standardization~\cite{Lawler2006}.

For our study, we use Project Gutenberg~\cite{Gutenberg1971} to obtain a large corpus of documents, and we extract a sequence of punctuation marks for each document in the corpus (see~\cref{sec:database}). Broadly, our goal is to investigate the following question: Do punctuation marks encode stylistic information about an author, a genre, or a time period? (Recall that we do not distinguish between the roles of authors and editors in a document, so our use of the word ``author'' is an expository shortcut.) Different writers have different writing styles (e.g., long versus short sentences, frequent versus sparse dialogue, and so on), and a writer's style can also evolve over time or differ across different types of works. It is plausible that an author's use of punctuation is --- consciously or unconsciously --- at least partly indicative of an idiosyncratic style, and we seek to explore the extent to which this is the case. Although there is a wealth of work that focuses on quantitative analysis of writing styles, punctuation marks and their (conscious or unconscious) stylistic footprints have largely been overlooked. Analysis of punctuation is also pertinent to ``prosody'', the study of the tune and rhythm of speech\footnote{An amusing illustration is the contrast between the Oxford comma, the Walken comma, and the Shatner comma. For one example, see \url{https://www.scoopnest.com/user/JournalistsLike/529351917986934784}.} and how these features contribute to meaning \cite{prosody}.

To the best of our knowledge, very few researchers have explored author recognition using only stylometric features that are punctuation-focused~\cite{Grieve2007, Chaski2001}. Additionally, the few existing works that include a punctuation-focused analysis used a very small author corpus ($40$ authors in~\cite{Grieve2007} and $5$ authors in~\cite{Chaski2001}) and focused on the frequency with which different punctuation marks occur (ignoring, e.g., the order in which they occur). In the present paper, we investigate author recognition using features that account for both the frequency and the order of punctuation marks in a corpus of \nbauthors\ authors and  \nbdocuments\ documents that we draw from the Project Gutenberg database (see~\cref{sec:authors}). Although Project Gutenberg is a popular database for the statistical analysis of language, most previous studies that have used it have considered only a small number of manually selected documents~\cite{Gerlach2018}.
We also use Project Gutenberg to explore genre recognition~\cite{Santini2004a, Santini2004b, Kessler1996, Chiang2015} from a punctuation perspective and stylochronometry~\cite{Can2004, Stamou2008, Parkes1992, Whissel1996, Jackson2002, Forsyth1999, Hughes2012} in~\cref{sec:genres} and~\cref{sec:time}, respectively.  There are not many studies of stylochronometry, and existing ones tend to be rather specific in nature (e.g., focused on particular authors, such as Shakespeare~\cite{Whissel1996} and band members from the Beatles~\cite{Jackson2002}, or on particular time frames)~\cite{Neal2017, Stamou2008}. Literary genre recognition (e.g., fiction, philosophy, etc.) has also received limited attention, and we are not aware of even a single study that has attempted genre recognition solely using punctuation. We wish to examine (1) whether punctuation is at all indicative of the style of an author, genre, or time period; and, if so, (2) the strength of stylistic signatures when one ignores words. In short, how much can one learn from punctuation alone?

Importantly, we do not seek to try to identify the best set of features for a given stylometric task, nor do we seek to conduct a thorough comparison of different methods for a given stylometric task. Instead, our goal is to give punctuation, an unsung hero of style, some overdue credit through an initial quantitative study of punctuation-focused stylometry. To do this, we focus on a small number of punctuation-related stylometric features and use this set of features to investigate questions in author recognition, genre recognition, and stylochronometry. To reiterate an important point, we do not account for an editor's effect on an author's style in our analysis, and it is important to interpret all of our findings with that caveat in mind. Given the supra-linguistic nature of punctuation and our reliance on punctuation-based features, one can perform an analysis like ours across different languages that use the same set of punctuation (e.g., across different translations). We offer a novel perspective on stylometry that we hope others will carry forward in their own punctuational pursuits, which include many exciting future directions.

Our paper proceeds as follows. We describe our data set (as well as our filtering and cleaning of it), punctuation-based features, and classification techniques in~\cref{sec:database}. We compare the use of punctuation across authors in~\cref{sec:authors}, across genres in~\cref{sec:genres}, and over time in~\cref{sec:time}. We conclude and offer directions for future work in~\cref{sec:conc}. The data set of punctuation sequences that we use in this paper is available at \url{https://dx.doi.org/10.5281/zenodo.3605100}, and the code that we use to analyze punctuation sequences is available at \url{https://github.com/alex-darmon/punctuation-stylometry}. 


\section{Data and methodology}
\label{sec:database}

\noindent \texttt{"This sentence has five words. Here are five more words. Five-word sentences 
are fine. But several together become monotonous. Listen to what is happening. 
The writing is getting boring. The sound of it drones. It's like a stuck record. 
The ear demands some variety. Now listen. I vary the sentence length, and I 
create music. Music. The writing sings. It has a pleasant rhythm, a lilt, a 
harmony. I use short sentences. And I use sentences of medium length. And sometimes, 
when I am certain the reader is rested, I will engage him with a sentence of 
considerable length, a sentence that burns with energy and builds with all the impetus 
of a crescendo, the roll of the drums, the crash of the cymbals \emdash{} sounds that say listen 
to this, it is important."
 \newline
\indent \indent \indent \indent \indent \indent \indent \indent \indent \indent \indent \indent	{\normalfont\emdash{}\;Gary Provost, \textit{100 Ways to Improve Your Writing}, 1985.}
}


\subsection{Data set} \label{sec:database1}

We use the API functionality of Project Gutenberg~\cite{Gutenberg1971} to obtain our document corpus and the natural-language-processing (NLP) library {\sc spaCy}~\cite{spacy2017} to extract a punctuation sequence from each document.\footnote{Many abbreviations, such as ``Dr.'' and ``Mr.'', are treated as words in {\sc spaCy}. Therefore, {\sc spaCy} does not count the periods in them as punctuation marks.} Using data from Project Gutenberg requires several filtering and cleaning steps before it is meaningful to perform statistical analysis~\cite{Gerlach2018}. We describe our steps below. 

\begin{figure}[t]
\centering
\includegraphics[width =7cm]{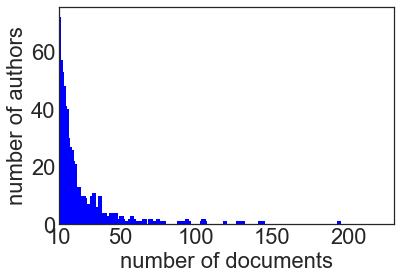}
\caption[]{Histogram of the number of documents per author in our corpus. 
}  
\label{distribution_books}
\end{figure}

We retain only documents that are written in English (a document's language is specified in metadata). We remove the author labels ``Various'', ``Anonymous'', and ``Unknown''. To try and mitigate, in an automated way, the issue of a document appearing more than once in our corpus (e.g., ``Tales and Novels of J. de La Fontaine -- Complete'', ``The Third Part of King Henry the Sixth'', ``Henry VI, Part 3'', ``The Complete Works of William Shakespeare'', and ``The History of Don Quixote, Volume 1, Complete''), we ensure that any given title appears only once, and we remove all documents with the word ``complete'' in the title.\footnote{It is still possible for a document to appear more than once in our corpus (e.g., ``The Third Part of King Henry the Sixth'' and ``Henry VI, Part 3''). We manually remove such duplicates when investigating specific authors over time (see~\cref{sec:time}).} 
(Note that the word ``anthology'' does not appear in any titles in our final corpus.) We also adjust some instances where a punctuation mark or a space appears incorrectly in the Project Gutenberg raw data (specifically, instances in which a double quotation appears as unicode or the spacing between words and punctuation marks is missing), and we remove any documents in which double quotations do not appear.\footnote{The latter may be legitimate documents, but we remove them to err on the side of caution.} Among the remaining documents, we retain only authors who have written at least 10 documents in our corpus. For each of these documents, we remove headers using the python function ``{\sc strip\_headers}'', which is available in Gutenberg's Python package. This yields a data set with \nbauthors\ authors and \nbdocuments\ documents. We show this final list of authors in~\cref{app:authorsgenres}. We show the distribution of documents per author in~\cref{distribution_books}. The documents in our corpus have various metadata, such as author birth year, author death year, document ``bookshelf'' (with at most one unique bookshelf per document), document subject (with multiple subjects possible per document), document language, and document rights. In some of our computational experiments, we use the following metadata: author birth year, author death year, and document ``bookshelf'' (which we term document ``genre'', as that is what it appears to represent). Gerlach and Font-Clos~\cite{Gerlach2018} pointed out recently that ``bookshelf'' may be better suited than ``subject'' for practical purposes such as text classification, because the former constitute broader categories and provide a unique assignment of labels to documents.

For each document, we extract a sequence of the following 10 punctuation marks: the period ``\;.\;''; the comma  ``\;,\;''; the colon ``\;:\;''; the semicolon ``\;;\;''; the left parenthesis ``\;(\;''; the right parenthesis ``\;)\;''; the question mark ``\;?\;''; the exclamation mark ``\;!\;''; double quotation marks, ``\;\;``\;\;'' and ``\;\;''\;\;'' (which are not differentiated consistently in Project Gutenberg's raw data); single quotation marks, ``\;\;`\;\;'' and ``\;\;'\;\;'' 
(which are also not differentiated consistently in Project Gutenberg's raw data), 
which we amalgamate with double quotation marks; and the ellipsis ``\; ... \;''. To promote a language-independent approach to punctuation (e.g., apostrophes in French can arise as required parts of words), we do not include apostrophes in our analysis. We also do not include hyphens, en dashes, or em dashes, as these are not differentiated consistently in Project Gutenberg's raw data and we find the choices among these marks in different documents --- standard rules of language be damned --- to be unreliable upon a visual inspection of some documents in our corpus.


\subsection{Features}
\label{sec:features}

Using standard terminology from the machine-learning literature, we use the word ``feature'' to refer to any quantitative characteristic of a document or set of documents. We compute six feature vectors for each document $k$ in our corpus to quantify the frequency with which punctuation marks occur, the order in which they occur, and the number of words that tend to occur between them. Specifically, we compute the following: 
\begin{itemize}
\item[(1)]{$\p{f}^{1,k}$, the frequency vector for punctuation marks in a given document $k$;}
\item[(2)]{$\p{f}^{2,k}$, an empirical approximation of the conditional probability of the successive occurrence of elements in an ordered pair of punctuation marks in document $k$;} 
\item[(3)]{$\p{f}^{3,k}$, an empirical approximation of the joint probability of the successive occurrence of elements in an ordered pair of punctuation marks in document $k$;} 
\item[(4)]{$\p{f}^{4,k}$, the frequency vector for sentence lengths in a given document $k$, where we consider the end of a sentence to be marked by a period, exclamation mark, question mark, or ellipsis;} 
\item[(5)]{$\p{f}^{5,k}$, the frequency vector for the number of words between successive punctuation marks in a given document $k$; and} 
\item[(6)]{$\p{f}^{6,k}$, the mean number of words between successive occurrences of the elements in an ordered pair of punctuation marks in document $k$.} 
\end{itemize}

We summarize these features in~\cref{features} and define each of these six features below. When appropriate, we suppress the superscript $k$ (which indexes
the document for which we compute a feature) from $\p{f}^{i,k}$ for ease of writing.

\begin{table}
\caption{Summary of the punctuation-sequence features that we study. See the text for details and mathematical formulas.}\label{features}
\begin{tabular}{ccc}
\hline\\ [-5.3ex]\hline
Feature & Description & Formula \\ \hline
$\p{f}^{1}$ & Punctuation-mark frequency & ~\eqref{PF} \\ \hline
$\p{f}^{2}$ & Conditional frequency of successive punctuation marks & ~\eqref{TM} \\ \hline
$\p{f}^{3}$ & Frequency of successive punctuation marks & ~\eqref{NTM}\\ \hline
$\p{f}^{4}$ & Sentence-length frequency & ~\eqref{SL}\\ \hline
$\p{f}^{5}$ & Frequency of number of words between successive punctuation marks & ~\eqref{WF}\\ \hline
$\p{f}^{6}$ & \begin{tabular}[c]{@{}c@{}}Mean number of words between successive occurrences\\ of the elements in ordered pairs of punctuation marks\end{tabular} & ~\eqref{WN}\\
\hline\\[-5.3ex]\hline
\end{tabular}
\end{table}

Let $\Theta=\{\theta_1,\ldots,\theta_{10}\}$ denote the (unordered) set of 10 punctuation marks (see~\cref{sec:database1}). Let $n$ denote the total number of documents in our corpus; and let $D_k=\{\theta_1^k,\ldots,\theta_{n_k}^k\}$, with $k \in \{1, \ldots, n\}$, denote the
sequence of $n_k$ punctuation marks in document $k$. As an example, consider the following quote by Ursula K. Le Guin (from an essay in her 2004 collection, \textit{The Wave in the Mind}):
\vspace{1em}

\texttt{
  I don't have a gun and I don't have even one wife and my sentences tend to go on and on and on, with all this syntax in them. Ernest Hemingway would have died rather than have syntax. Or semicolons. I use a whole lot of half-assed semicolons; there was one of them just now; that was a semicolon after "semicolons," and another one after "now."
 \vspace{1em}}

The sequence $D_k$ for this quote is $\{$, \,\textbar\, . \,\textbar\, . \,\textbar\, . \,\textbar\, ; \,\textbar\, ; \,\textbar\, ``  \,\textbar\, , \,\textbar\, ''  \,\textbar\, ``  \,\textbar\, . \,\textbar\, ''$\}$,\footnote{Because there can be commas in the elements of some of the sets and sequences that we consider (e.g., the sequence $D_k$), we use vertical lines instead of commas to separate elements in sets and sequences with
punctuation marks to avoid confusion. }
and there are $n_k  = 12$ punctuation marks. From $D_k$, we can calculate $\p{f}^{1,k}, \p{f}^{2,k}$, and $\p{f}^{3,k}$.

\begin{figure}[t]
\centering
\subfigure[Exclamation mark]{
\includegraphics[width = 3.1cm]{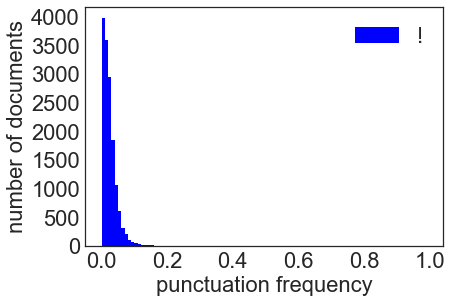}}
\subfigure[Quotation mark]{
\includegraphics[width = 3.1cm]{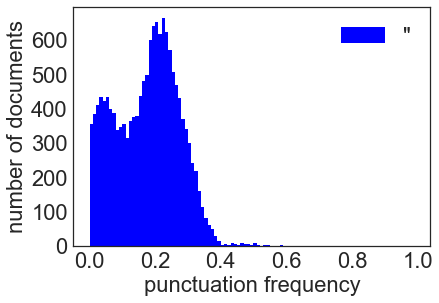}}
\subfigure[Left parenthesis]{
\includegraphics[width = 3.1cm]{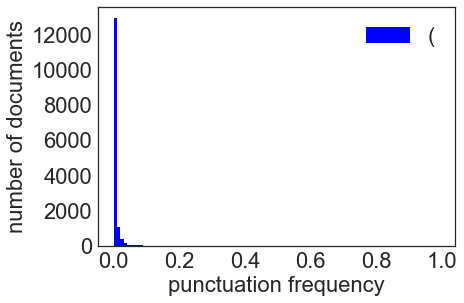}}
\subfigure[Right parenthesis]{
\includegraphics[width = 3.1cm]{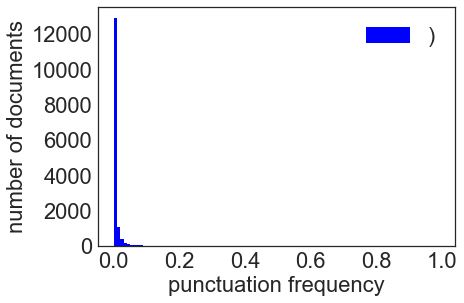}}

\subfigure[Comma]{
\includegraphics[width = 3.1cm]{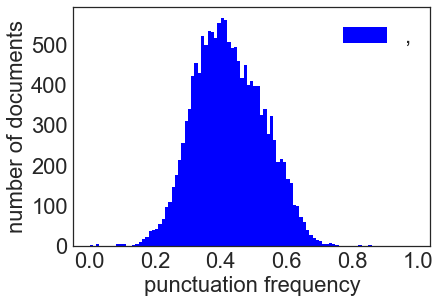}}
\subfigure[Period]{
\includegraphics[width = 3.1cm]{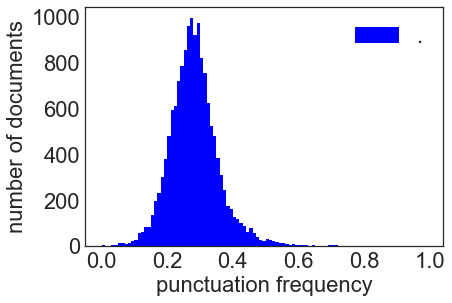}}
\subfigure[Colon]{
\includegraphics[width = 3.1cm]{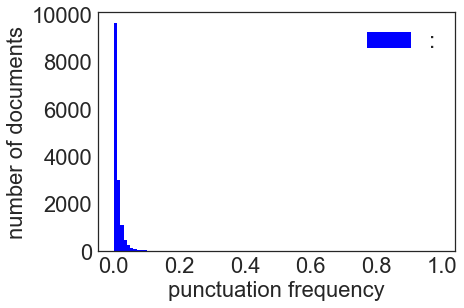}}

\subfigure[Semicolon]{
\includegraphics[width = 3.1cm]{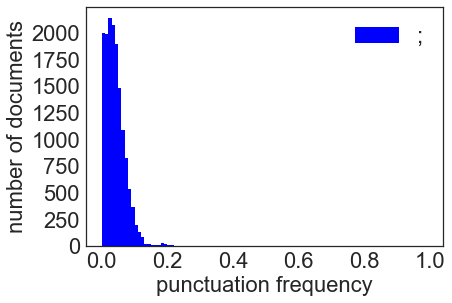}}
\subfigure[Question mark]{
\includegraphics[width = 3.1cm]{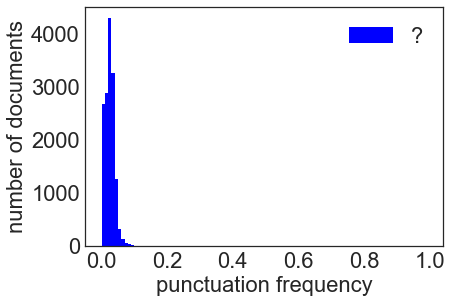}}
\subfigure[Ellipsis]{
\includegraphics[width = 3.1cm]{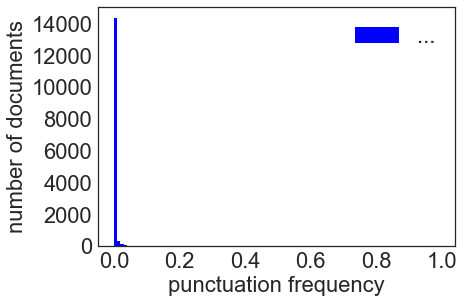}}
\caption{
Histogram of punctuation-mark frequencies of the documents in our corpus. The horizontal axis of each panel gives the frequency of a punctuation mark binned by $0.01$, and the vertical axis of each panel gives the total number of documents in our corpus with a punctuation-mark frequency in the bin. That is, the first bar of a panel for punctuation mark $\theta_i$ indicates the number of documents in our corpus for which $0 \leq f^{1,k}_i < 0.01$, the second bar indicates the number of documents in our corpus for which $0.01 \leq f^{1,k}_i < 0.02$, and so on. In descending order, the means (rounded to the third decimal) of each set $\{f^{1,k}_i\,, k=1, \ldots , n\}$ (which we use to construct our plot for $\theta_i$) are $0.024$ (exclamation mark), $0.175$ (apostrophe), $0.006$ (left parenthesis), $0.006$ (right parenthesis), $0.425$ (comma), $0.283$ (period), $0.013$ (colon), $0.041$ (semicolon), $0.025$ (question mark), and $0.002$ (ellipsis). These numbers imply that, on average, $42.5\%$ of the punctuation of a document in our corpus consists of commas, $28.3\%$ of it consists of periods, $4.1\%$ of it consists of semicolons, and so on.
}
\label{feature_dist}
\end{figure}

We determine each entry of $\p{f}^{1,k}$ from the number of times that the
associated punctuation mark appears in a document, relative to the total
number of punctuation marks in a document:
\begin{equation}
	f^{1,k}_i = \frac{\lvert \{\theta_l^k \in D_k \,\vert\, \theta_l^k = \theta_i \}\rvert}{n_k}\,.
\label{PF}
\end{equation}
The feature $\p{f}^{1,k}$ induces a discrete probability distribution on the set of punctuation marks for each document in our corpus (i.e., $\sum_{i=1}^{\vert\Theta\vert}f^{1,k}_i = 1$ for all $k$) and is independent of the order of the punctuation marks. For the Le Guin quote, 
\begin{equation*}
\p{f}^{1} =  \begin{bmatrix}

\textbf{!} & \textbf{\texttt{"}} & \textbf{(} & \textbf{)} & \textbf{,} & \textbf{.} & \textbf{:} & \textbf{;} & \textbf{?} & \textbf{...}\\ 
0 &  \frac{1}{3} & 0 & 0 & \frac{1}{6} & \frac{1}{3} & 0 & \frac{1}{6} & 0 & 0
        \end{bmatrix}\,,
\end{equation*}
where the second row indicates the elements of the vector and the first row indicates the corresponding punctuation marks. (Recall from~\cref{sec:database1} that we amalgamate opening and closing double and single quotation marks into a single punctuation mark, so that entry refers to the appearance of either of those two marks.) An alternative is to consider the frequency of punctuation marks relative to the number of characters or words in a document~\cite{Grieve2007}. 
  In \cref{feature_dist}, we 
  show the histograms of punctuation-mark frequencies (which are given by $\p{f}^{1}$) across all documents in our corpus. 
  These plots give an idea of the overall usage of each punctuation mark in our corpus. For instance, we see that commas and periods are (unsurprisingly) the most common punctuation marks in the corpus documents. We also observe that the comma frequency varies more across documents than the period frequency. Another observation is that there appear to be two peaks in quotation-mark frequency: a lower peak at about $0.1$ (with a height of approximately $450$ documents) and a higher peak at about $0.25$ (with a height of approximately $650$ documents). No other punctuation mark has more than one    noticeable peak; this may suggest that one can cluster documents in our corpus into two sets whose characteristic feature is how often they use quotation marks. 

To compute $\p{f}^{2,k}$ and $\p{f}^{3,k}$, we consider a categorical Markov chain on the sequence of punctuation marks and associate each punctuation mark with a state of the Markov chain. We first need two types of transition matrices. We calculate the matrix \mbox{$\p{P}^{k}\in[0,1]^{\vert\Theta\vert\times\vert\Theta\vert}$} from the number of times that elements in an ordered pair of punctuation marks occur successively in a document, relative to the number of times that the first punctuation mark in this pair occurs in the document: 
\begin{equation} \label{TM}
	P^k_{ij} = \frac{\vert\{\theta_l^k \in D_k\vert \theta_l^k = \theta_i \text{ and } \theta_{l+1}^{k} = \theta_j\}\vert}{\lvert \{\theta_l^k \in D_k\,\vert\, \theta_l^k = \theta_i \}\rvert}\,,\quad \text{with}\quad  \sum_{j}P_{ij}^k = 1\,.
      \end{equation}  
When a punctuation mark $\theta_i$ does not appear in a document, we set all entries in the corresponding row to $0$. We calculate the matrix \mbox{$\widetilde{\p{P}}^k\in[0,1]^{\vert\Theta\vert\times\vert\Theta\vert}$} from the number of times that elements in an ordered pair of successive punctuation marks occur in a document, relative to the total number of punctuation marks in the document:
\begin{equation} \label{NTM} \widetilde{P}^k_{ij} =
        \frac{\vert\{\theta_l^k \in D_k\vert \theta_l^k = \theta_i \text{ and
          } \theta_{l+1}^{k} = \theta_j \}\vert}{n_k}\,, \quad \text{with}\quad 
        \sum_{i,j}\widetilde{P}_{ij}^k = 1\,.
\end{equation}
Note that $\widetilde{P}^k_{ij} = P^k_{ij}f^{1,k}_i $. 

The transition matrix $\p{P}^{k}$ is an estimate of the conditional probability
of observing punctuation mark $\theta_j$ after punctuation mark $\theta_i$ in
document $k$, and the transition matrix $\widetilde{\p{P}}^k$ is an estimate of the joint
probability of observing the punctuation marks $\theta_i$ and $\theta_j$ in succession in document $k$. 
The relationship $\widetilde{P}^k_{ij} = P^k_{ij}f^{1,k}_i$ ensures that rare
(respectively, frequent) events are given less (respectively, more) weight in
$\p{\widetilde{P}}$ than in $\p{P}$. For example, if an author seldom uses
the ellipsis ``...'' in a document, the few ways in which it was used (which, arguably, are not
representative of authorial style) are assigned high probabilities in $\p{P}$ but low probabilities in
$\p{\widetilde{P}}$. For the Le Guin quote, $\p{P}$ and $\p{\widetilde{P}}$ are
\begin{equation*}
   \p{P} =  \begin{bmatrix}
   \textbf{!} & \textbf{\texttt{"}} & \textbf{(} & \textbf{)} & \textbf{,} & \textbf{.} & \textbf{:} & \textbf{;} & \textbf{?} & \textbf{...}\\ \hline
   0 & 0 & 0 & 0 & 0 & 0 & 0 & 0 & 0 & 0\\
    0 & \frac{1}{3} & 0 & 0 & \frac{1}{3} & \frac{1}{3} & 0 & 0 & 0 & 0\\
     0 & 0 & 0 & 0 & 0 & 0 & 0 & 0 & 0 & 0\\
      0 & 0 & 0 & 0 & 0 & 0 & 0 & 0 & 0 & 0\\
       0 & \frac{1}{2} & 0 & 0 & 0 & \frac{1}{2} & 0 & 0 & 0 & 0\\
        0 & \frac{1}{4} & 0 & 0 & 0 & \frac{1}{2} & 0 & \frac{1}{4} & 0 & 0\\
         0 & 0 & 0 & 0 & 0 & 0 & 0 & 0 & 0 & 0\\
          0 & \frac{1}{2} & 0 & 0 & 0 & 0 & 0 & \frac{1}{2} & 0 & 0\\
           0 & 0 & 0 & 0 & 0 & 0 & 0 & 0 & 0 & 0\\
            0 & 0 & 0 & 0 & 0 & 0 & 0 & 0 & 0 & 0\\
  \end{bmatrix}\,, \quad
  \p{\widetilde{P}} =  \begin{bmatrix}
 \textbf{!} & \textbf{\texttt{"}} & \textbf{(} & \textbf{)} & \textbf{,} & \textbf{.} & \textbf{:} & \textbf{;} & \textbf{?} & \textbf{...}\\ \hline
   0 & 0 & 0 & 0 & 0 & 0 & 0 & 0 & 0 & 0\\
    0 & \frac{1}{9} & 0 & 0 & \frac{1}{9} & \frac{1}{9} & 0 & 0 & 0 & 0\\
     0 & 0 & 0 & 0 & 0 & 0 & 0 & 0 & 0 & 0\\
      0 & 0 & 0 & 0 & 0 & 0 & 0 & 0 & 0 & 0\\
       0 & \frac{1}{12} & 0 & 0 & 0 & \frac{1}{12} & 0 & 0 & 0 & 0\\
        0 & \frac{1}{12} & 0 & 0 & 0 & \frac{1}{6} & 0 & \frac{1}{12} & 0 & 0\\
         0 & 0 & 0 & 0 & 0 & 0 & 0 & 0 & 0 & 0\\
          0 & \frac{1}{12} & 0 & 0 & 0 & 0 & 0 & \frac{1}{12} & 0 & 0\\
           0 & 0 & 0 & 0 & 0 & 0 & 0 & 0 & 0 & 0\\
            0 & 0 & 0 & 0 & 0 & 0 & 0 & 0 & 0 & 0\\
  \end{bmatrix}\,,
\end{equation*}
where the first row of each matrix indicates the corresponding punctuation mark. Observe that $\widetilde{P}_{56} < \widetilde{P}_{66}$, even though these entries are equal in $\p{P}$, because two successive periods occur more frequently than a period followed by a comma in Le Guin's quote.  

We obtain $\p{f}^{2,k}$ and $\p{f}^{3,k}$ by ``flattening''
(i.e., concatenating the rows of) the matrices $\p{P}^{k}$ and
$\p{\widetilde{P}}^{k}$, respectively. 
For example, we obtain $\p{f}^2$ for the Le Guin quote by appending the rows of $\p{P}$ in order and one after the other.
The feature $\p{f}^{3,k}$ induces a joint
probability distribution on the space of ordered punctuation
pairs. In contrast to $\p{f}^{1,k}$, the features $\p{f}^{2,k}$ and $\p{f}^{3,k}$ depend on the order in which punctuation marks occur in a document. As we will see in~\cref{sec:authors}, the feature $\p{f}^{3,k}$ is very effective at distinguishing different authors. 
We account for order with a one-step lag in $\p{f}^{2,k}$ and $\p{f}^{3,k}$ (i.e., each state depends only on the previous state). One can generalize these features to account for memory or \quoting{long-range correlations}~\cite{Ebeling1993}. For example, the probability of closing a parenthesis increases after it has been opened.
 
The features $\p{f}^{4,k}$, $\p{f}^{5,k}$, and $\p{f}^{6,k}$ account for the number of words that occur between punctuation marks. Let $D_k^w=\{w_0^k, w_1^k,\ldots,w_{n_k-1}^k\}$ denote the number of words that occur between successive punctuation marks in $D_k$, with $w_0^k$ equal to the number of words before the first punctuation mark. Therefore, $w_1^k$ is the number of words between punctuation marks $\theta_1^k$ and $\theta_2^k$, and so on. 
The sequence $D_k^w$ for Le Guin's comment is $\{25, 6,  9,  2, 9, 7, 5, 1, 0, 4,  1,  0\}$, where we count ``don't'' as two words and we also count ``half-assed'' as two words. The minimum number of words that can occur
between successive punctuation marks is $0$, and we cap the maximum number of
words that can occur between successive punctuation marks at $n_s = 40$ and
the number of words in a sentence at $n_S = 200$. 
Fewer than 0.05\,\% of the sentences in our corpus exceed $n_S = 200$ words; similarly, the $n_s = 40$ cap is exceeded by fewer than 0.05\,\% of the strings between successive punctuation marks.

The entries of the feature $\p{f}^{4,k}\in[0,1]^{n_S\times 1}$, which
quantifies the frequency of sentence lengths, are
\begin{equation}   \label{SL}
	  f^{4,k}_i = \frac{\vert\{w_l^k\in D_k^w\,\vert\, w_l^k = i \text{ and } \theta_l,\theta_{l+1}\in\{.\,\vert\,...\,\vert\, ! \,\vert\, ?\}\,\}\vert}{n_k}\,.
\end{equation}
In the Le Guin quote, there are four sentences, with lengths $31$, $9$, $2$, and $27$ (in sequential order). The feature $\p{f}^{4,k}$, an $n_S\times 1$ vector with $n_S = 200$, thus has the value $1/4$ in the $9^{\text{th}}$, $2^{\text{nd}}$, $27^{\text{th}}$, and $31^{\text{st}}$ positions and the value $0$ in all other entries. One can also consider other measures of sentence length (e.g., the number of characters, instead of the number of words)~\cite{Vieira2018}. 

The entries of the feature $\p{f}^{5,k}\in[0,1]^{n_s\times 1}$, which quantifies the frequency of the number of words between successive punctuation marks, are 
\begin{equation}   \label{WF}
	  f^{5,k}_i = \frac{\vert\{w_l^k\in D_k^w\,\vert\, w_l^k = i \}\vert}{n_k}\,\,.
\end{equation}
In the Le Guin quote, recall that $D_k^w = \{25, 6,  9,  2, 9, 7, 5, 1, 0, 4,  1,  0\}$ (which includes $9$ unique integers), so the  $n_s\times 1$ vector (with $n_s = 40$, as mentioned above) $\p{f}^{5}$ has $9$ nonzero entries. For example, $\p{f}^{5}_1 = 2/12$ (because $0$ occurs twice out of $n_k = 12$ total punctuation marks) and $\p{f}^{5}_4 = 0$ (because $3$ never occurs out of $n_k = 12$ possible times). 

The features $\p{f}^{4,k}$ and $\p{f}^{5,k}$ induce discrete probability distributions on the number of words in sentences and the number of words between successive punctuation marks, respectively. The expectation of the feature $\p{f}^{5,k}$ quantifies the ``rate of punctuation'' and is equal to the total number of words, relative to the total number of punctuation marks: 
\begin{equation} \label{punctuation_rate}
	\mathbb{E}\left[f^{5,k}\right] = \sum_{i=0}^{n_s} i \times f^{5,k}_i 
= \frac{1}{n_k} \times \sum_{i=0}^{n_s} i \times \left\vert\{w_l^k\in D_k^w\,\vert\, w_l^k = i \}\right\vert = \frac{\vert D_k^w \vert}{n_k}\,.
\end{equation}
The feature $\p{f}^{5,k}$ tracks word-count frequency between successive punctuation marks, without distinguishing between different punctuation marks. 

With $\p{f}^{6,k}$, we compute the mean number of words between successive occurrences of the elements in  ordered pairs of punctuation marks using a matrix $\p{W}^k\in[0,n_s]^{\vert \Theta\vert \times \vert\Theta\vert}$ with entries
\begin{equation}   \label{WN}
	 W_{ij}^k = \langle{\{w_l^k\in D_k^w\,\vert\, \theta_l = \theta_i  \,\text{  and  }\, \theta_{l+1}=\theta_j\}\rangle}\,,
\end{equation}
where $\langle \, \cdot \, \rangle$ denotes the sample mean of a set. The matrix for the Le Guin excerpt is 
\begin{equation*}
   \p{W} =  \begin{bmatrix}
   \textbf{!} & \textbf{\texttt{"}} & \textbf{(} & \textbf{)} & \textbf{,} & \textbf{.} & \textbf{:} & \textbf{;} & \textbf{?} & \textbf{...}\\\hline
   0 & 0 & 0 & 0 & 0 & 0 & 0 & 0 & 0 & 0\\
    0 & 4 & 0 & 0 & 1 & 1 & 0 & 0 & 0 & 0\\
     0 & 0 & 0 & 0 & 0 & 0 & 0 & 0 & 0 & 0\\
      0 & 0 & 0 & 0 & 0 & 0 & 0 & 0 & 0 & 0\\
       0 & 0 & 0 & 0 & 0 & 0 & 0 & 0 & 0 & 0\\
        0 & 0 & 0 & 0 & 0 & 5.5 & 0 & 9 & 0 & 0\\
         0 & 0 & 0 & 0 & 0 & 0 & 0 & 0 & 0 & 0\\
          0 & 5 & 0 & 0 & 0 & 0 & 0 & 7 & 0 & 0\\
           0 & 0 & 0 & 0 & 0 & 0 & 0 & 0 & 0 & 0\\
            0 & 0 & 0 & 0 & 0 & 0 & 0 & 0 & 0 & 0\\
  \end{bmatrix}\,.
 \end{equation*}
 We obtain $\p{f}^{6,k}$ by flattening the matrix $\p{W}^k$ by concatenating its rows. As variants of this feature, one need not require that punctuation-mark occurrences are successive, and one can subsequently compute the number of words or even the number of (other) punctuation marks between the elements of an ordered pair of punctuation marks.

\begin{figure}[t]
\centering
\subfigure[$\p{f}^1$ \textit{King Lear}, W. Shakespeare]{
\includegraphics[width = 3cm]{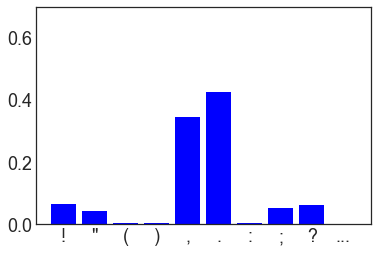}}
\subfigure[$\p{f}^1$ \textit{Hamlet}, W. Shakespeare]{
\includegraphics[width = 3cm]{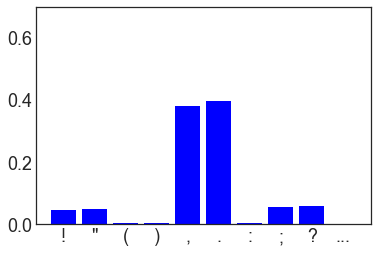}}
\subfigure[$\p{f}^1$ \textit{The History of Mr. Polly}, H. G. Wells]{
\includegraphics[width = 3cm]{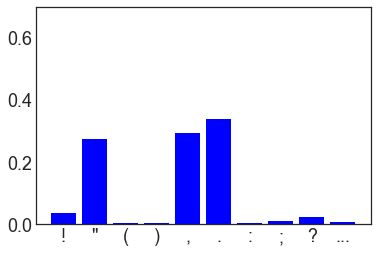}}
\subfigure[$\p{f}^1$ \textit{The Wheels of Chance}, H. G. Wells]{
\includegraphics[width = 3cm]{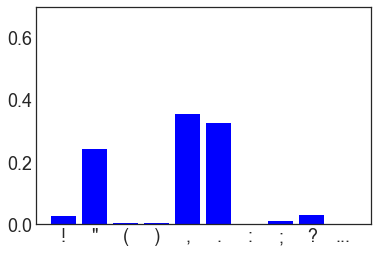}}
\subfigure[$\p{f}^5$ \textit{Hamlet}, W. Shakespeare]{
\includegraphics[width = 3cm]{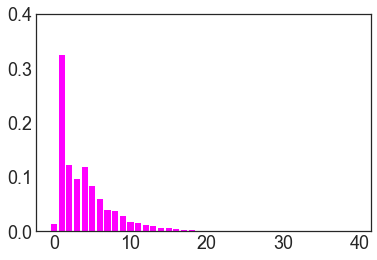}}
\subfigure[$\p{f}^5$ \textit{Hamlet}, W. Shakespeare]{
\includegraphics[width = 3cm]{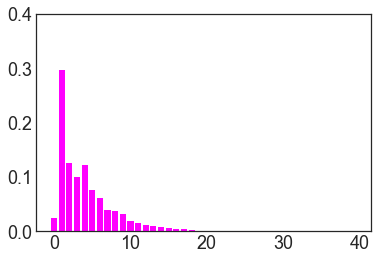}}
\subfigure[$\p{f}^5$ \textit{The History of Mr. Polly}, H. G. Wells]{
\includegraphics[width = 3cm]{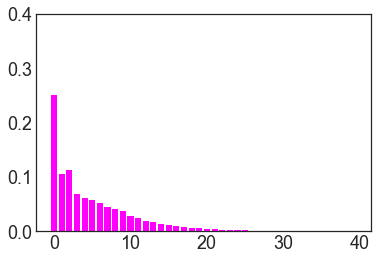}}
\subfigure[$\p{f}^5$ \textit{The Wheels of Chance}, H. G. Wells]{
\includegraphics[width = 3cm]{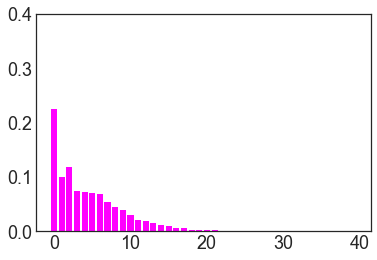}}
\caption{(a,b,c,d) Histograms of punctuation-mark frequency ($\p{f}^1$) and (e,f,g,h) the number of words that occur between successive punctuation marks ($\p{f}^5$) for two documents by William Shakespeare and two documents by Herbert George Wells.
 }  
\label{feature_freq}
\end{figure}

In the rest of our paper, we focus on the six features $\p{f}^{1},\, \ldots,\, \p{f}^{6}$. We show example histograms of $\p{f}^1$ (punctuation frequency) and $\p{f}^5$ (mean number of words between successive punctuation marks) for some documents by the same authors in~\cref{feature_freq}.

\subsection{Kullback--Leibler divergence}
\label{KL}

To quantify the similarity between two discrete distributions (e.g., between
the features $\p{f}^1, \p{f}^3, \p{f}^4$, and $\p{f}^5$ from different documents), we use Kullback--Leibler (KL)
divergence~\cite{Kullback1951}, an information-theoretic measure that is related to Shannon entropy and ideas from maximum-likelihood theory. KL divergence and variants of it have been used in prior research on author recognition~\cite{Neal2017, Arun2009, Zhao2006}. One can also consider other similarity measures, such as chi-square distance~\cite{Neal2017} and Jensen--Shannon divergence~\cite{Lin1991, Altmann2017, Gerlach2016}.

Consider a random variable $X$ with a discrete, finite support
$x\in\mathcal{X}$; and let $\p{p}\in[0,1]^{\vert\mathcal{X}\vert\times 1}$
and $\p{q}\in[0,1]^{\vert\mathcal{X}\vert\times 1}$ be two probability
distributions for $X$ that we assume are absolutely continuous with respect to each other. Broadly speaking, KL divergence quantifies how close a probability
distribution \mbox{$\p{p} = \{p_i\}$} is to a candidate distribution $\p{q}=\{q_i\}$, where $p_i$ (respectively, $q_i$) denotes the probability that $X$ takes the value $i$ when it is distributed according to $\p{p}$ (respectively, $\p{q}$)~\cite{Cover1991}. The KL divergence between the probability distributions $\p{p}$ and $\p{q}$ is defined as 
\begin{equation}    \label{KL-equ}
	   d_{\mathrm{KL}}(p, q) = \sum_{i=1}^{\vert\mathcal{X}\vert} p_i \log\,\left( \frac{p_i}{q_i}\right)
\end{equation}
and satisfies four important properties:
\begin{enumerate}
\item{$d_{\mathrm{KL}}(\p{p},\p{q}) \geq 0$\,;}
\item{$d_{\mathrm{KL}}(\p{p},\p{q}) = 0$ if and only if $p_i = q_i$ for all $i$\,;}
\item{$d_{\mathrm{KL}}(.\,,\,.)$ is asymmetric in its arguments; and}
\item{$d_{\mathrm{KL}}(\p{p},\p{q}) = H(\p{p},\p{q}) - H(\p{p})$\,,}
\end{enumerate}
where $H(\p{p})=\sum_ip_i\log p_i$ denotes the Shannon entropy of $\p{p}$ and $H(\p{p}, \p{q})$ denotes the Shannon entropy of the joint distribution of $\p{p}$ and $\p{q}$~\cite{Shannon1948, Lesne2014}. Entropy quantifies the ``unevenness'' of a probability distribution. It represents the mean information that is required to specify an outcome of a random variable, given its
probability distribution. It achieves its minimum value $0$ for a constant random variable (e.g.,
$p_1=1$ and $p_i = 0$ for $i\neq 1$) and its maximum value
$\log(\vert\mathcal{X}\vert)$ for a uniform distribution.  In some sense,
$d_{\mathrm{KL}}(\p{p},\p{q})$ measures the ``unevenness'' of the joint distribution
of $\p{p}$ and $\p{q}$ relative to the distribution of $\p{p}$. One can also derive KL divergence from likelihood theory. 
In particular, one can show that, as the number of samples from the discrete random variable $\mathcal{X}$ tends to infinity, KL divergence measures the mean likelihood of observing data with the distribution $\p{p}$  if the distribution $\p{q}$ actually generated the data~\cite{Shlens2014, Duda2001}. 

To adjust for cases in which $\p{p}$ and $\p{q}$ are not absolutely continuous with respect to each other (e.g., one document has one or more ellipses, but another does not, resulting in unequal supports), we remove any frequency component that corresponds to a punctuation mark that is not in the common support and then distribute the weight of the removed frequency uniformly across the other frequencies. For example, suppose that $p_1\neq 0$ but $q_1 = 0$. We then define $\p{\widetilde{p}}$ such that $\p{\widetilde{p}} = \{p_i/(1 - p_1)\,, \,\,\, i\neq 1\}$ and compute $d_{\mathrm{KL}}(\widetilde{\p{p}}, \p{q})$.


\subsection{Classification models}
\label{class_models}

We describe the two classification approaches that we use for author recognition (see~\cref{sec:author_rec}) and genre recognition (see~\cref{sec:genres_rec}). Much of the existing classification work on author recognition uses machine-learning classifiers (e.g., support vector machines or neural networks) or similarity-based classification techniques (e.g., using KL divergence)~\cite{Neal2017, Stamatatos2009}. We use neural networks and similarity-based classification with KL divergence for both author and genre classification. Following standard practice, we split the $n$ documents in our data set into into a training set and a testing set. Broadly speaking, a training set calibrates a classification model (e.g., to ``feed'' a neural network and adjust its parameters), and one then uses a testing set to evaluate the accuracy of a calibrated model. We ensure that all authors or genres (i.e., all ``classes'') that appear in the testing set also appear in the training corpus; this is known as ``closed-set attribution'' and is common practice in author recognition~\cite{Stamatatos2009, Neal2017}. For a given data set, we place 80\% of the documents in the training set and the remaining 20\% of documents in the testing set. (A training:testing ratio of 80:20 is a common choice.)
A given data set is sometimes the entire corpus (i.e., \nbdocuments\ documents and \nbauthors\ authors), and it is sometimes a subset of it. In our summary tables (see \cref{sec:author_rec} and~\cref{sec:genres_rec}), we explicitly specify the sizes of the training and testing sets of our experiments.


\subsubsection{Similarity-based classification}
\label{multiclass_NN}

We label our $p$ classes by $c_1, c_2, \ldots, c_p$ (recall that these can correspond to authors or genres), and we denote the set of training documents for class $c_j$ by  $\mathcal{D}_j$. For each class $c_j$, we define a class-level feature $\p{f}^{l,c_j}$, with $l\in\{1,\ldots,6\}$ and $j\in\{1,\ldots, p\}$, by averaging the features across the training documents in that class. That is, the $i^{\text{th}}$ entry of  $\p{f}^{l,c_j}$ is 
\begin{equation}
	f_{i}^{l,c_j} = \frac{1}{\vert \mathcal{D}_j\vert}\sum_{k\in \mathcal{D}_j}f^{l,k}_i \,, 
\end{equation}
where $l\in\{1,\ldots,6\}$ and we use the features $\p{f}^{1,k}, \p{f}^{2,k}, \ldots, \p{f}^{6,k}$ from~\cref{sec:features}. This yields a set $\phi^{k} = \{\p{f}^{1,k}, \ldots,\p{f}^{6,k}\}$ of features for each document and a set $\phi^{c_j} = \{\p{f}^{1,c_j}, \ldots,\p{f}^{6,c_j}\}$ of features for each class.

To determine which class is ``most similar'' to a document $k$ in our testing set, we solve the following minimization problem:
\begin{equation}
	    \text{argmin}_{j\in\{1,\ldots,p\}}d(\phi^k,\phi^{c_j})\,,
\end{equation}
for some choice of similarity measure $d(\,.\,,\,.\,)$. In our numerical experiments of~\cref{sec:authors}, we use the KL-divergence similarity measure $d_{\mathrm{KL}}$ to define $d(\,.\,,\,.\,)$ as 
\begin{equation}
	    d(\phi^k,\phi^{c_j}) = \text{argmin}_{l\in\mathcal{L}} d_{\mathrm{KL}}(\p{f}^{l,c_j},\p{f}^{l,k})\,,
\end{equation}
where we restrict the set of features to those that induce discrete probability distributions and consider each feature individually (i.e., $\mathcal{L} = \{1\}$, $\mathcal{L} = \{3\}$, $\mathcal{L} = \{4\}$, or $\mathcal{L} = \{5\}$). 


\subsubsection{Neural networks}
\label{multiclass_NN2}

We use feedforward neural networks with the standard backpropagation
algorithm as a machine-learning classifier \cite{nn_kjell}. A neural network
uses the features of a training set to automatically infer rules for
recognizing the classes of a testing set by adjusting the weights of each
``neuron'' using a stochastic gradient-descent-based learning algorithm. 
In contrast with neural networks for classical NLP classification, where it is standard to use word embeddings and employ convolutional or recurrent neural networks \cite{convoNLP} to ensure that input vectors have equal lengths, we have already defined our features such that they have equal length. It thus suffices for us to use feedforward neural networks. The input vector that corresponds to each document is a
concatenation of the six features (or a subset
thereof) in \cref{sec:features}, and the output is a probability vector, which one can interpret as the likelihood that a given document belongs to a given class. We assign each document in our testing set to the class with highest probability.


\subsection{Model evaluation}
\label{evaluation}

For each test of a classification model, we consider a data set with a fixed number of classes (e.g.,  \nbauthors\ classes if we perform author recognition on all authors in our corpus), a uniformly-randomly sampled training set (80\% of the data set), and a testing set (the remaining 20\% of the data set). We measure ``accuracy'' as the ratio of correctly assigned documents relative to the total number of documents in a testing set. 
For each test of a classification model, we report two quantities: (1) the accuracy of the classification model on the testing set;
and (2) the accuracy of a baseline classifier on the testing set, which we obtain by assigning each document in the testing set to each class with a probability that is proportional to the class's size in the training set.


\section{Case study: Author analysis}
\label{sec:authors}

\texttt{\noindent "It is almost always a 
greater pleasure to come across a semicolon than a period. The period tells 
you that that is that;
if you didn't get all the meaning you wanted or 
expected, anyway you got all the writer intended to parcel out and now you 
have to move along. 
But with a semicolon there you get a pleasant little 
feeling of expectancy; there is more to come; to read on; it will get clearer."
 \newline
\indent \indent \indent \indent \indent \indent \indent \indent \indent \indent \indent \indent	{\normalfont \emdash{}\;Thomas Lewis, \textit{Notes on Punctuation}, 1979}}


\subsection{Consistency}

\begin{figure}[t!]
\centering
\subfigure[\textit{Sharing Her Crime}, M. A. Fleming]{
\includegraphics[width = 3cm]{figs2/Sharing_Her_Crime_A_Novel_heatmap.png}}
\subfigure[\textit{The Actress' Daughter}, M. A. Fleming]{
\includegraphics[width = 3cm]{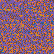}}
\hspace{4mm}
\subfigure[\textit{King Lear}, W. Shakespeare]{
\includegraphics[width = 3cm]{figs2/King_Lear_heatmap.png}}
\subfigure[\textit{Hamlet}, W. Shakespeare]{
\includegraphics[width = 3cm]{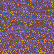}}

\centering\includegraphics[width = 4cm]{figs2/color_bar_heatmap_seqpun.png}

\subfigure[\textit{The History of Mr. Polly}, H. G. Wells]{
\includegraphics[width = 3cm]{figs2/The_History_of_Mr_Polly_heatmap.png}}
\subfigure[\textit{The Wheels of Chance}, H. G. Wells]{
\includegraphics[width = 3cm]{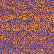}}
\caption[]{Sequences of successive punctuation marks that we extract from documents by (a,b) May Agnes Fleming, (c,d) William Shakespeare, and (e,f) Herbert George Wells. We map each punctuation mark to a distinct color. We cap the length of the punctuation sequence at $3000$ entries, which start at the midpoint of the punctuation sequence of the corresponding document.
}
\label{heatmaps_punctuation_marks}
\end{figure}

We explore punctuation sequences of a few authors to gain some insight into whether certain authors have more distinguishable punctuation styles than others. (Once again, recall our cautionary note that we do not distinguish between the roles of authors and editors for the documents in our corpus.) In~\cref{heatmaps_punctuation_marks}, we show (augmenting~\cref{fig:raw_sequence}) raw sequences of punctuation marks for two books by each of the following three authors: May Agnes Fleming, William Shakespeare, and Herbert George (H. G.) Wells. We observe for this document sample that, visually, one can correctly guess which documents were written by the same author based only on the sequences of punctuation marks. This striking possibility was illustrated previously in A.~J. Calhoun's blog entry \cite{Calhoun2016blog}, which motivated our research. From ~\cref{heatmaps_punctuation_marks}, we see that Wells appears to use noticeably more quotation marks than the other two authors. We also observe that Shakespeare appears to use more periods than Wells. These observations are consistent with the histograms in~\cref{feature_freq} (where we also observe that Shakespeare appears to use more exclamation marks and question marks than Wells), which we compute from the entire documents, so our observations from the samples in~\cref{heatmaps_punctuation_marks} appear to hold throughout those documents.

\begin{figure}[t!]
\centering
\subfigure[\textit{Sharing Her Crime} and \textit{The Actress' Daughter}]{
\includegraphics[width = 3.5cm]{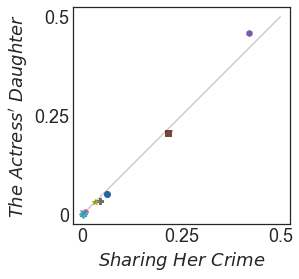}}
\subfigure[\textit{The History of Mr. Polly} and \textit{The Wheels of Chance}]{
\includegraphics[width = 3.5cm]{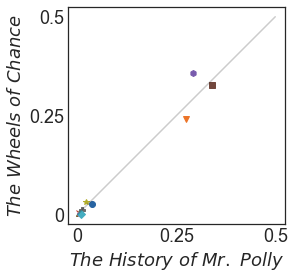}}
\subfigure[\textit{King Lear} and \textit{Hamlet}]{
\includegraphics[width = 3.5cm]{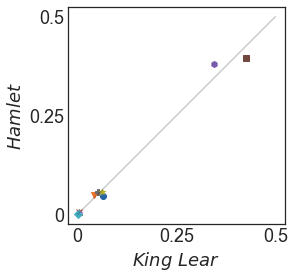}}
\centering\includegraphics[width = 7cm]{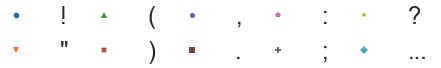}

\subfigure[\textit{The Actress' Daughter} and \textit{King Lear}]{
\includegraphics[width = 3.5cm]{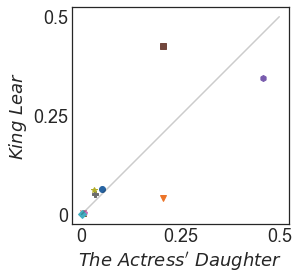}}
\subfigure[\textit{Hamlet} and \textit{The History of Mr.Polly}]{
\includegraphics[width = 3.5cm]{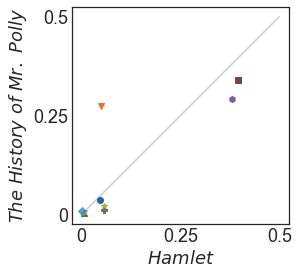}}
\subfigure[\textit{The Wheels of Chance} and \textit{Sharing Her Crime}]{
\includegraphics[width = 3.5cm]{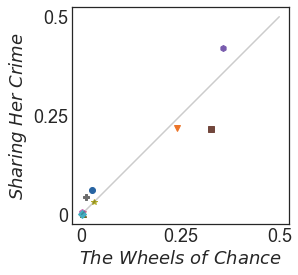}}
\caption[]{Scatter plots of frequency vectors (i.e., $\p{f^{1}}$) of punctuation marks to compare books from the same author: (a) \textit{Sharing Her Crime} and \textit{The Actress' Daughter} by May Agnes Fleming,
(c) \textit{King Lear} and \textit{Hamlet} by William Shakespeare,
and (e) \textit{The History of Mr. Polly} and \textit{The Wheels of Chance} by H. G. Wells.
Scatter plots of frequency vectors of punctuation marks to compare books from different authors: 
(b) \textit{The Actress' Daughter} and \textit{King Lear},
(d) \textit{Hamlet} and \textit{The History of Mr. Polly},
and (f) \textit{The Wheels of Chance} and \textit{Sharing Her Crime}. We represent each punctuation mark by a colored marker, with coordinates given by the punctuation frequencies in a vector that is associated to each document. The gray line represents the identity function. More similar frequency vectors correspond to dots that are closer to the gray line.
}
\label{scatter_plot_freq_pun}
\end{figure}

In~\cref{scatter_plot_freq_pun}, we plot examples of the punctuation
frequency (i.e., $\p{f^{1}}$) of one document versus that of another document by the same author (top row) and a document by a different author (bottom row). We base these plots on the ``rank order'' plots in~\cite{Yang2003}, who used such plots to illustrate the top-ranking words in various texts. In our plots, any punctuation mark (which we represent by a colored marker) that has the same frequency in both documents lies on the gray diagonal line. Any marker above (respectively, below) the gray line signifies that it is used more (respectively, less) frequently by the author on the vertical axis (respectively, horizontal axis). In these examples, we see for documents by the same author that the markers tend to be closer to the gray line than for documents by different authors. In~\cref{scatter_plot_freq_pun}(d), for example, we observe that Fleming used more quotation marks and commas in \textit{The Actress' Daughter} than Shakespeare did in \textit{King Lear}, whereas Shakespeare used more periods in \textit{King Lear} than Fleming did in \textit{The Actress' Daughter}. One can make similar observations about panels (e) and (f) of~\cref{scatter_plot_freq_pun}. These observations are consistent with those of~\cref{feature_freq} and \cref{heatmaps_punctuation_marks}.

\begin{figure}[t]
\centering
\subfigure[$\p{f}^1$: 10 authors]{
\includegraphics[width = 3.1cm]{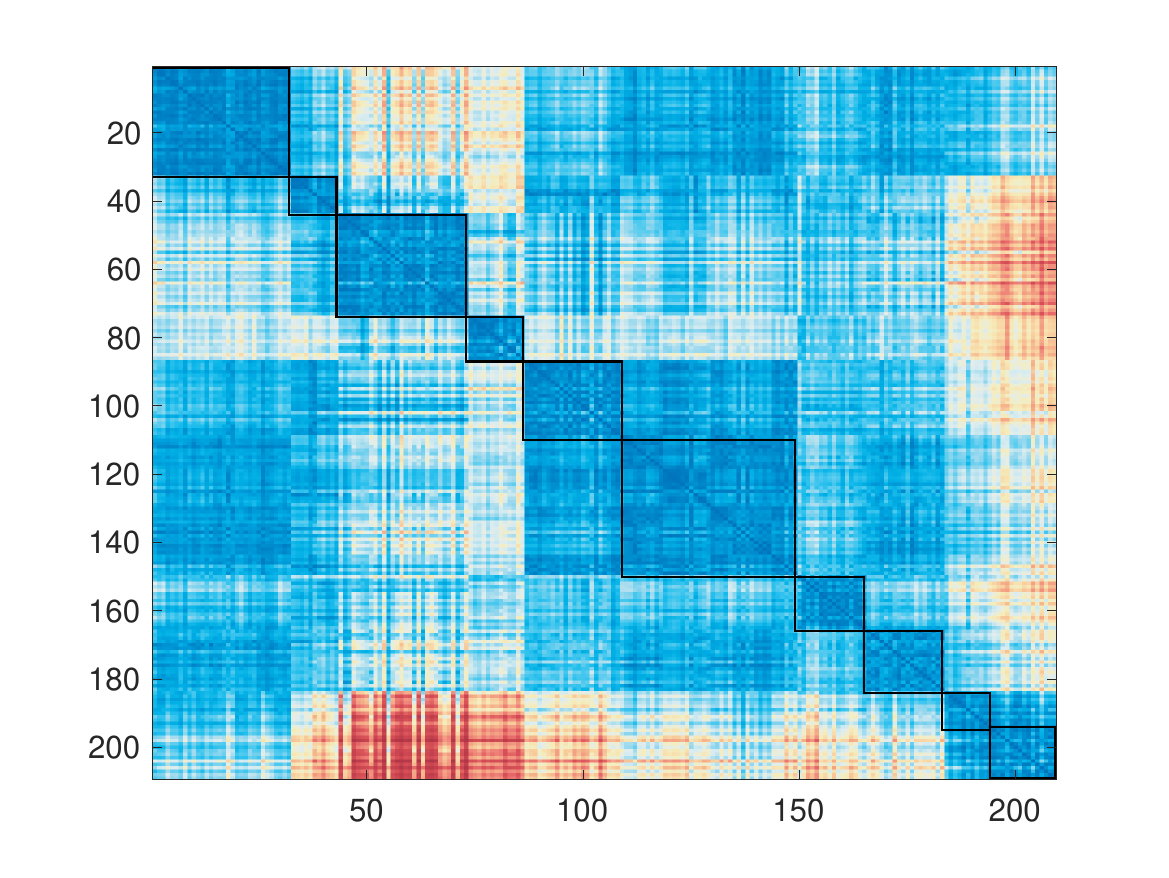}}
\subfigure[$\p{f}^3$: 10 authors]{
\includegraphics[width = 3.1cm]{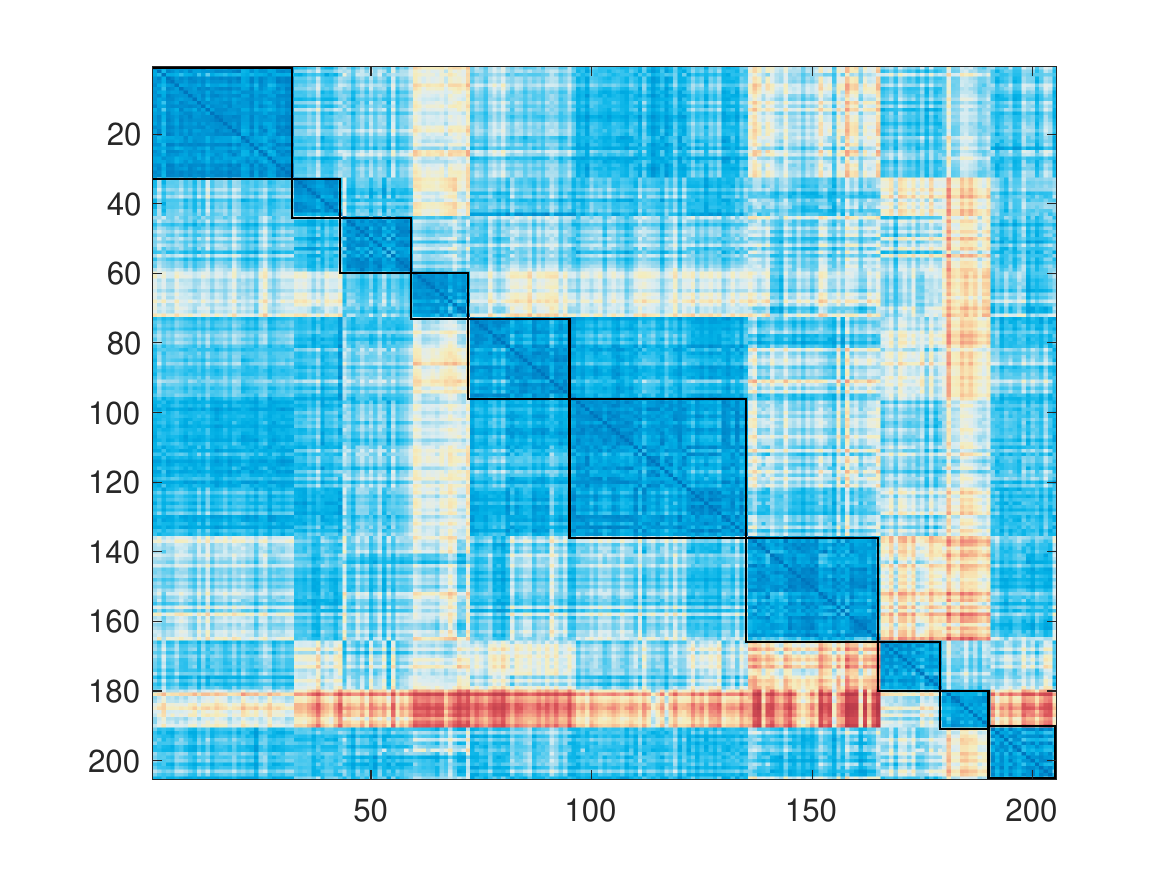}}
\subfigure[$\p{f}^4$: 10 authors]{
\includegraphics[width = 3.1cm]{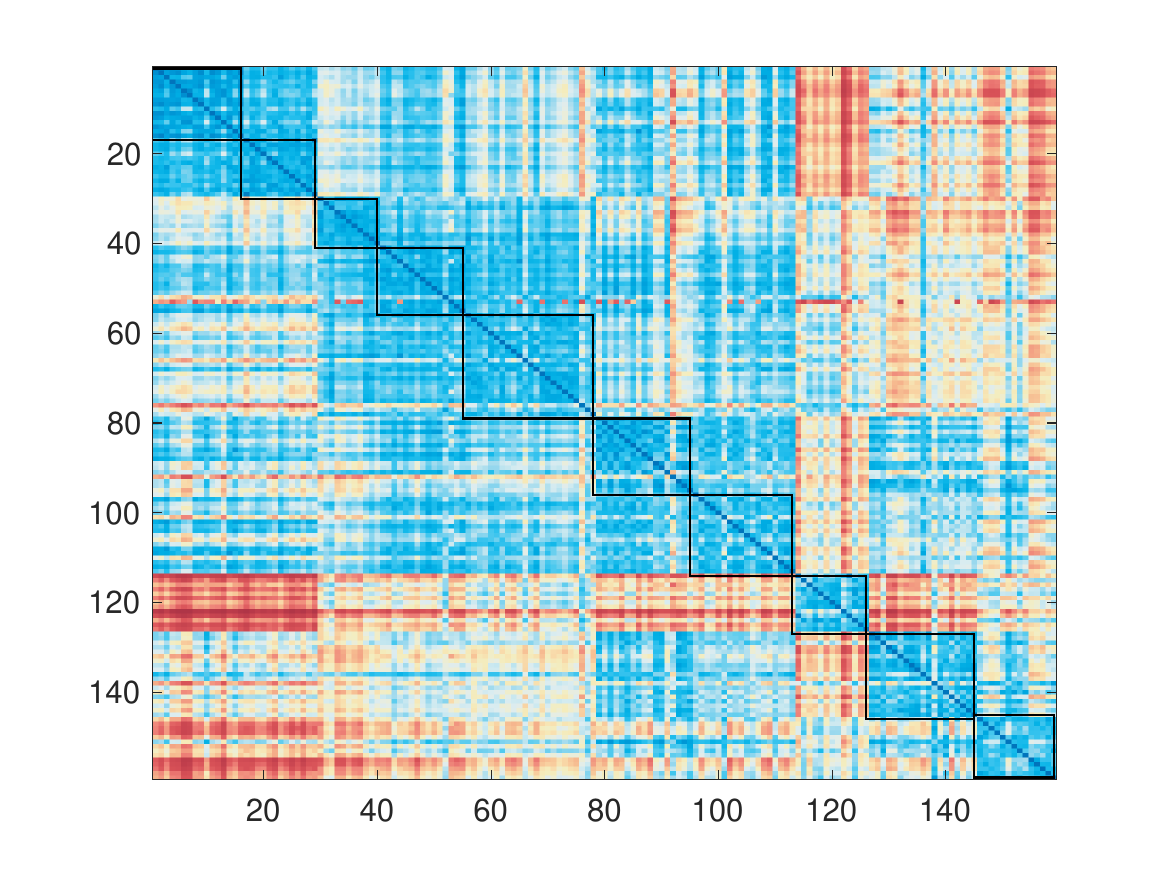}}
\subfigure[$\p{f}^5$: 10 authors]{
\includegraphics[width = 3.1cm]{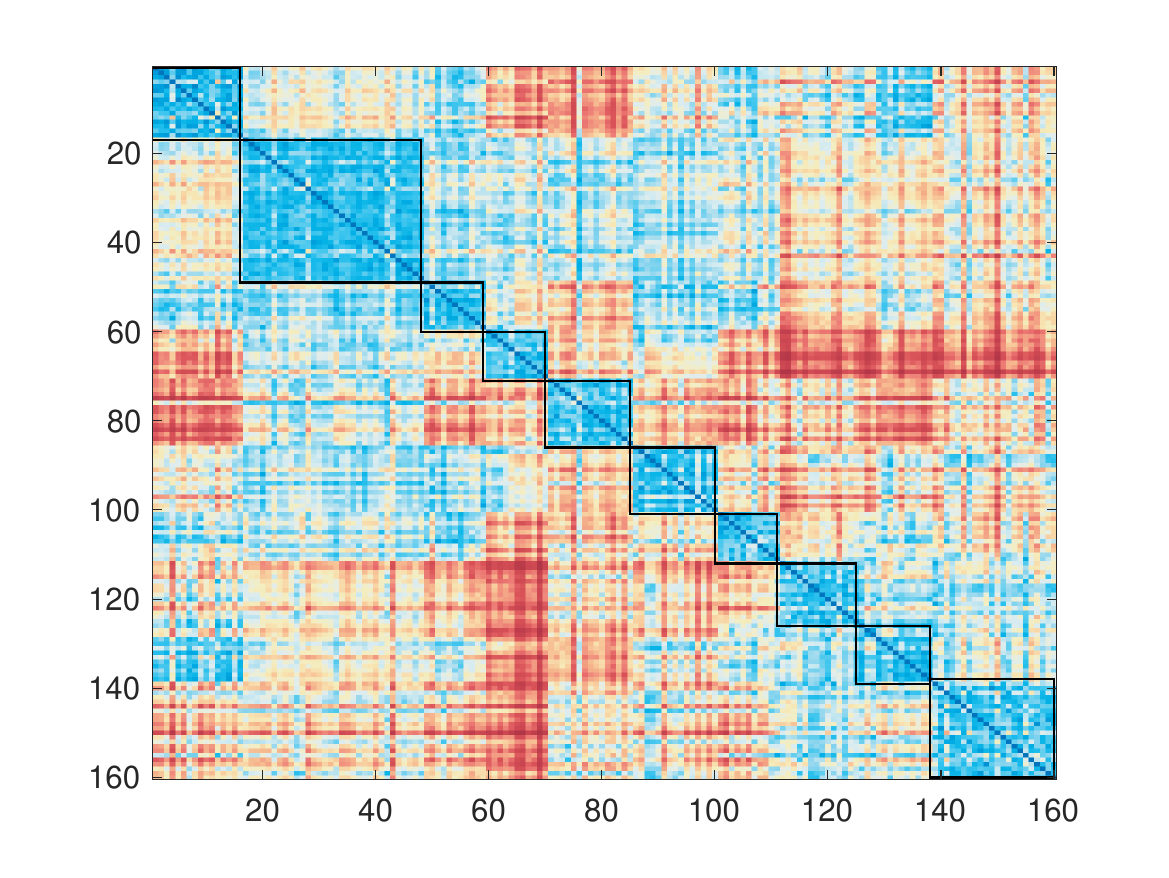}}

\subfigure[$\p{f}^1$: 50 authors]{
\includegraphics[width = 3.1cm]{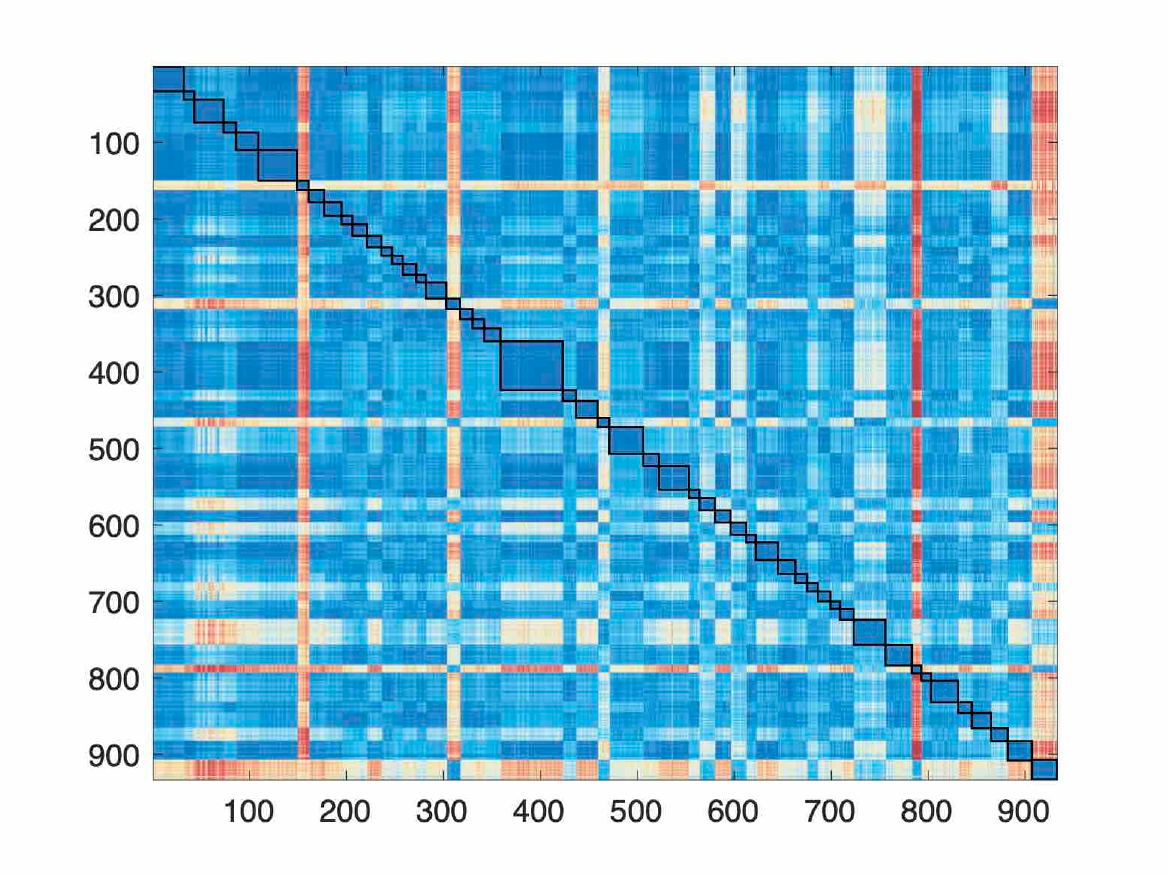}}
\subfigure[$\p{f}^3$: 50 authors]{
\includegraphics[width = 3.1cm]{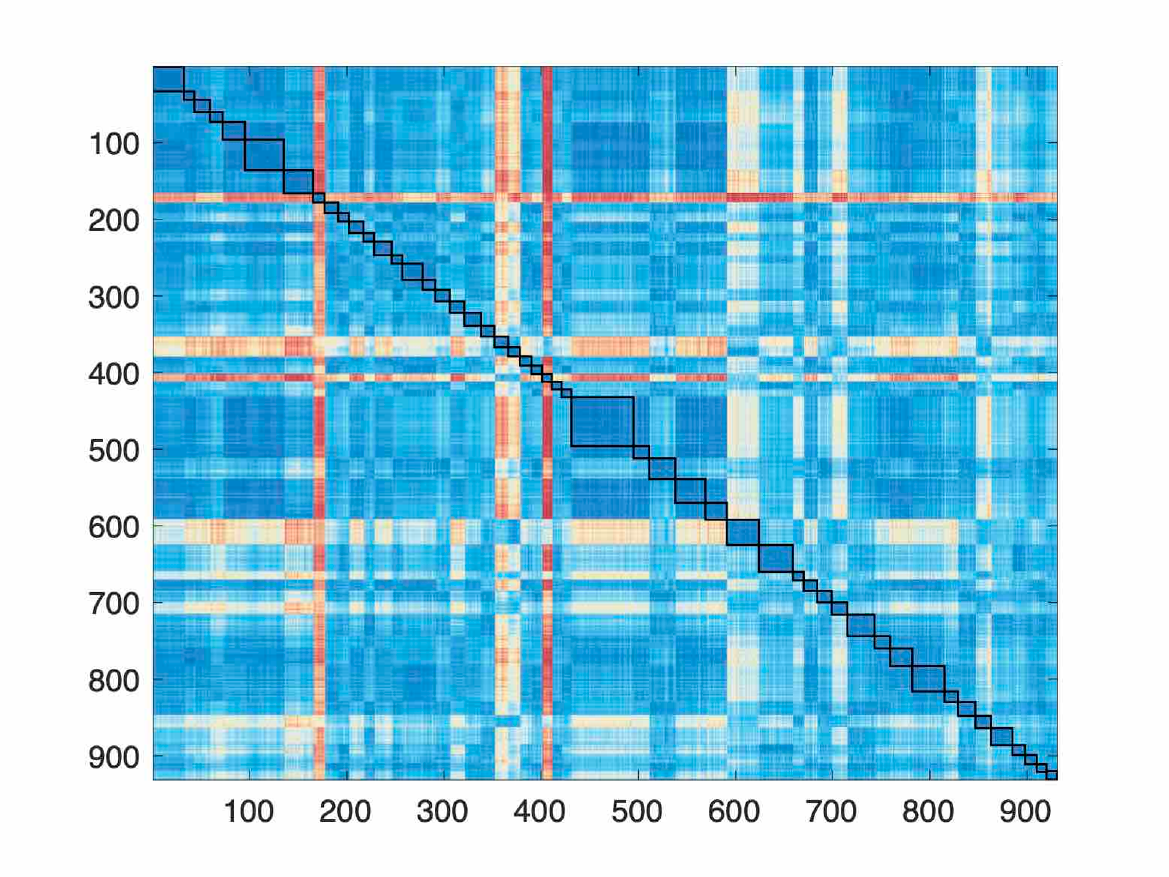}}
\subfigure[$\p{f}^4$: 50 authors]{
\includegraphics[width = 3.1cm]{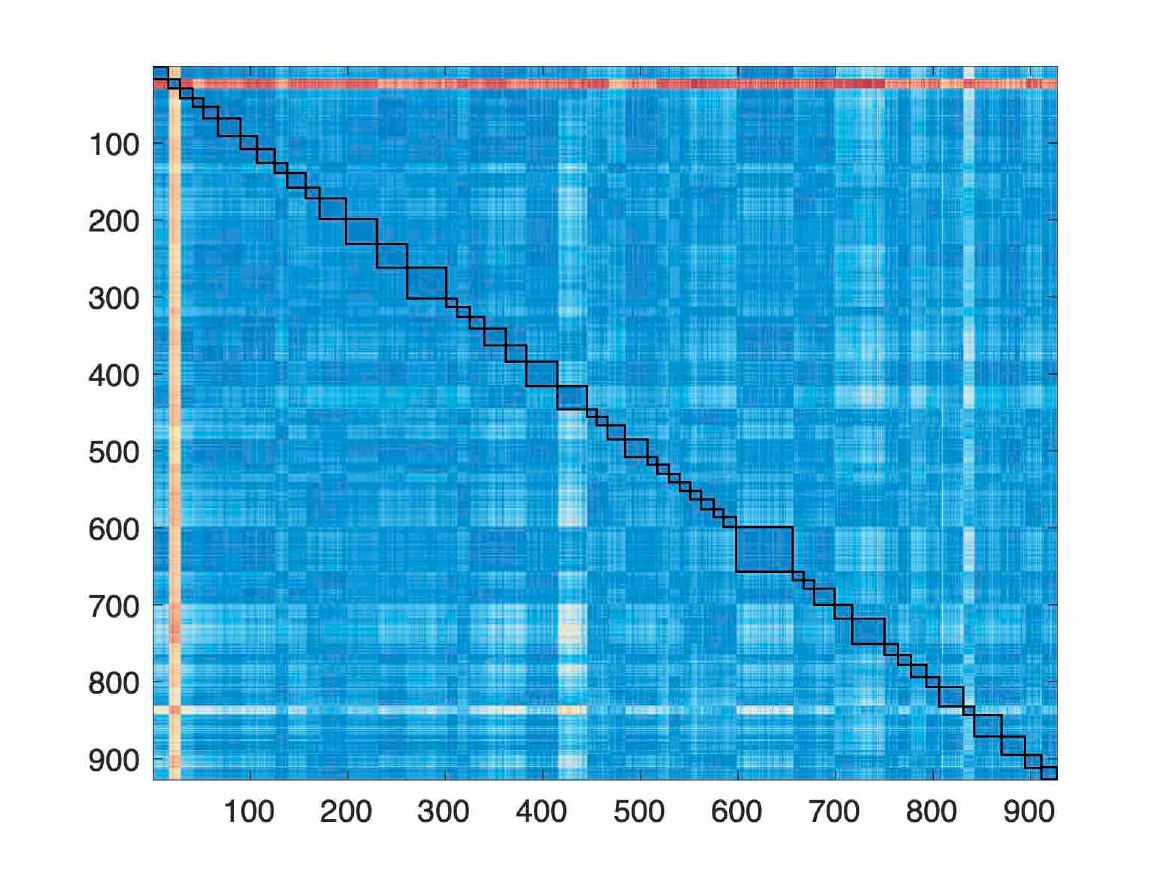}}
\subfigure[$\p{f}^5$: 50 authors]{
\includegraphics[width = 3.1cm]{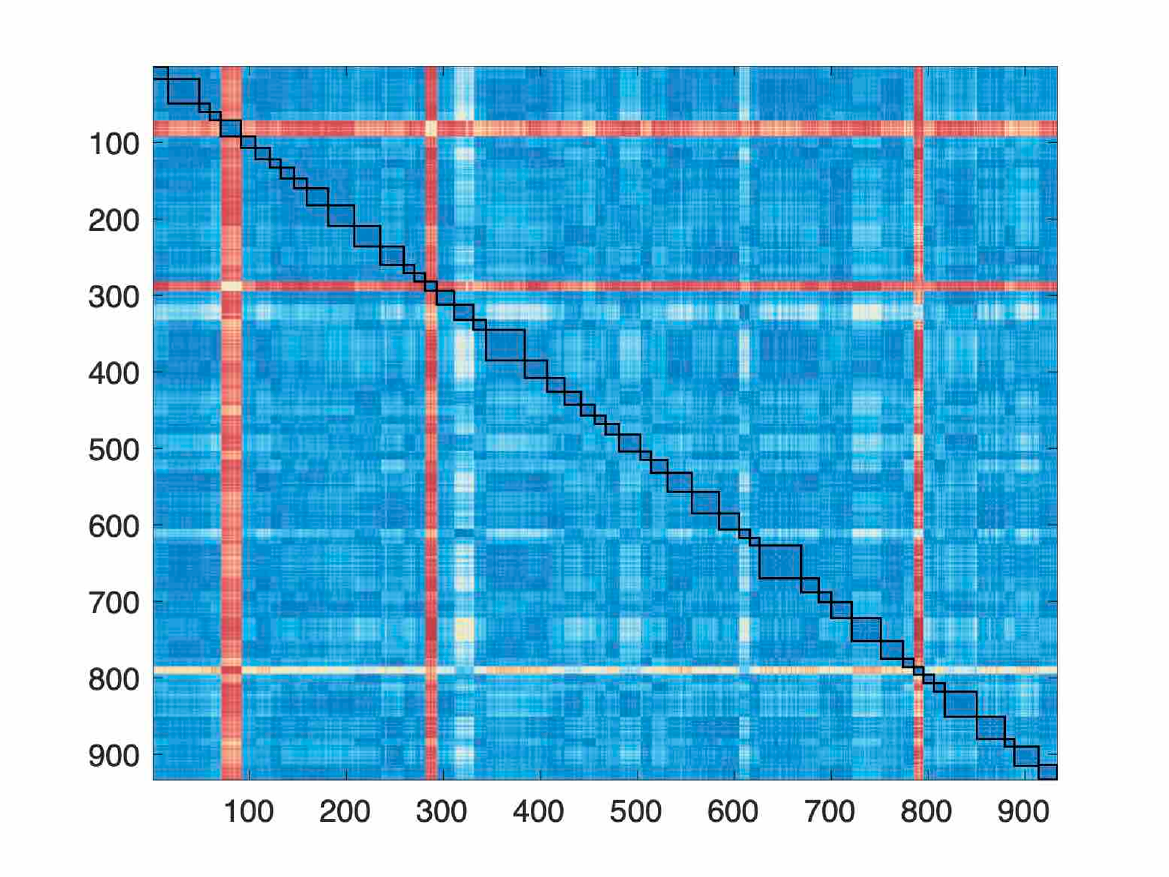}}

\caption[]{Heat maps showing KL divergence between the features (a,e) $\p{f}^1$, (b,f) $\p{f}^3$, (c,g) $\p{f}^4$, and (d,h) $\p{f}^5$ for different sets of authors. We show the 10 most-consistent (see the main text for our notion of ``consistency'') authors for each feature in the top row and the fifty most-consistent authors for each feature in the bottom row. The diagonal blocks that we enclose in black indicate documents by the same author. Authors can differ across panels, because author consistency can differ across features. The colors scale (nonlinearly) from dark blue (corresponding to a KL divergence of $0$) to dark red (corresponding to the maximum value of KL divergence in the underlying matrix). For ease of exposition, we suppress color bars (they span the interval $[0,3.35]$), given that the purpose of this figure is to illustrate the presence and/or absence of high-level structure. When determining the 10 most-consistent authors, we exclude the author ``United States. Warren Commission'' (see row 7 in~\cref{author-table}) in panels (a--c) and we exclude the author ``United States. Central Intelligence Agency'' (see row 116 in~\cref{author-table}) in panel (d). In each case, the we replace them with the next most-consistent author and proceed from there. Works by these two authors consist primarily of court testimonies or lists of definitions and facts (with rigid styles); they manifested as pronounced dark-red stripes that masked salient block structures.
 }
\label{heatmaps_KL}
\end{figure}

Our illustrations in~\cref{heatmaps_punctuation_marks} and~\cref{scatter_plot_freq_pun} use a very small number of documents by only a few authors. To quantify the ``consistency'' of an author across all documents by that author in our corpus, we use KL divergence. 

In~\cref{heatmaps_KL}, we show heat maps of KL divergence between discrete probability distributions induced by the feature vectors $\p{f}^1$, $\p{f}^3$, $\p{f}^4$, and $\p{f}^5$. We define the \todefine{consistency} of an author relative to a feature as the mean KL divergence for that feature computed across all pairs of documents by that author. That is, 
\begin{equation}     \label{consistency}
   	 \mathcal{C}_{\p{f}^i}(a) = \frac{2}{\vert \mathcal{D}_a -1\vert \vert\mathcal{D}_a\vert}\sum_{k,k'\in\mathcal{D}_a}d_{\mathrm{KL}}(\p{f}^{i,k},\p{f}^{i,k'})\,,
\end{equation}
where $a$ denotes an author in our corpus and $\mathcal{D}_a$ is the set of documents by author $a$. For each feature in~\cref{heatmaps_KL}, we show the 10 (respectively, 50) most-consistent authors in the top row (respectively, bottom row). Diagonal blocks with black outlines correspond to documents by the same author. Although there appears to be greater similarity within diagonal blocks than between them for several of the authors, it is difficult to interpret the heat maps when there are many authors (and it becomes increasingly difficult as one considers progressively more authors).

\begin{figure}[t]
\centering
\subfigure[feature $\p{f}^1$]{
\includegraphics[width = 3cm]{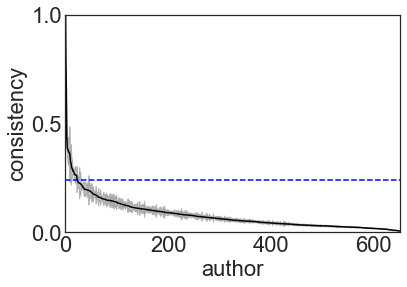}}
\subfigure[feature $\p{f}^3$]{
\includegraphics[width = 3cm]{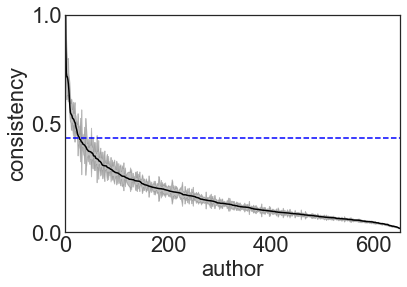}}
\subfigure[feature $\p{f}^4$]{
\includegraphics[width = 3cm]{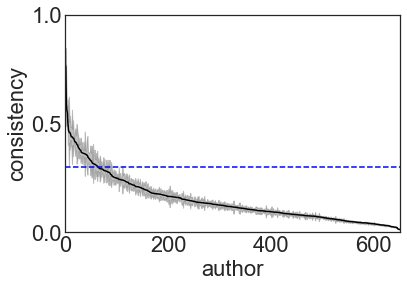}}
\subfigure[feature $\p{f}^5$]{
\includegraphics[width = 3cm]{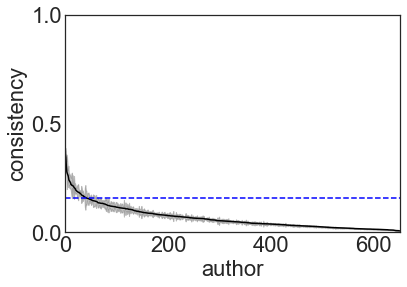}}
\centering\includegraphics[width = 5cm]{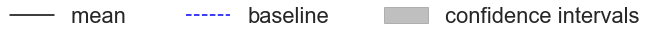}
\caption[]{Evaluations of author consistency. In each panel, we show author consistency~\eqref{consistency} for the features (a) $\p{f}^1$, (b) $\p{f}^3$, (c) $\p{f}^4$, and (d) $\p{f}^5$ using a solid black curve. In gray, we plot confidence intervals of KL divergence across pairs of documents for each author. To compute the confidence intervals, we assume that the KL divergence values across pairs of distinct documents for each author are normally distributed. There are at least $10$ documents by each author in our corpus (see~\cref{sec:database1}), so the number of KL values across pairs of distinct documents by a given author is at least $90$. The dotted blue line indicates a consistency baseline, which we obtain by choosing, uniformly at random, $1000$ ordered pairs of documents by distinct authors and computing the mean KL divergence between the features of these document pairs. 
}
\label{consistency_plots}
\end{figure}

In~\cref{consistency_plots}, we show author consistency in our entire corpus for the feature vectors $\p{f}^1$, $\p{f}^3$, $\p{f}^4$, and $\p{f}^5$. In each panel, we show a baseline (in blue), which we obtain by choosing, uniformly at random, $1000$ ordered pairs of documents by distinct authors and computing the mean KL divergence between the features of these document pairs. One pair is a single element of an off-diagonal block of a matrix like those in~\cref{heatmaps_KL}.  

We order each panel from the least-consistent author to the most-consistent author. Authors can differ across panels, because the consistency measure~\eqref{consistency} is a feature-dependent quantity. We observe in all panels of~\cref{consistency_plots} that most authors are more consistent on average than the baseline. (The black curve lies below the blue horizontal line for most authors.) The differences between authors relative to the baseline are most pronounced for the feature $\p{f}^3$ (see~\cref{features}). This suggests that $\p{f}^3$ may carry more information than our other five features about an author's idiosyncratic style. We come back to this observation in \cref{sec:author_rec}.

\begin{figure}[t]
\centering
\subfigure[feature $\p{f}^1$]{
\includegraphics[width = 3cm]{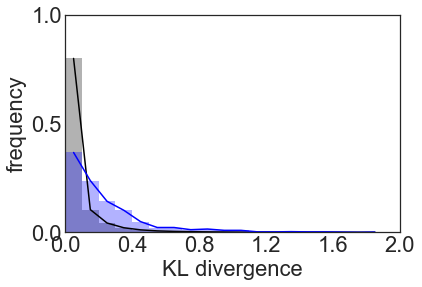}}
\subfigure[feature $\p{f}^3$]{
\includegraphics[width = 3cm]{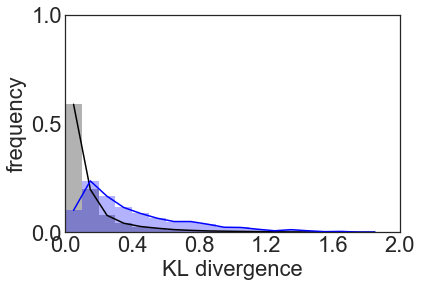}}
\subfigure[feature $\p{f}^4$]{
\includegraphics[width = 3cm]{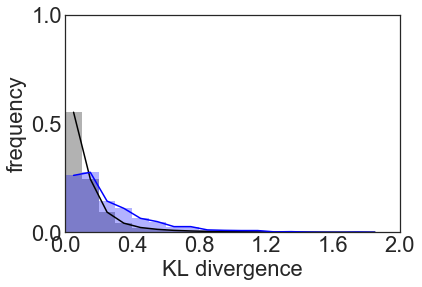}}
\subfigure[feature $\p{f}^5$]{
\includegraphics[width = 3cm]{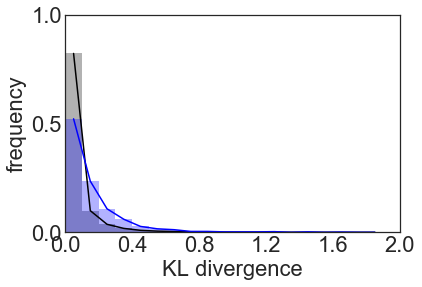}}

\caption[]{Distributions of KL divergence for authors. In each panel, we show the distributions of KL divergence between all pairs of documents in the corpus by the same author (in black) and between $1000$ ordered pairs of documents by distinct authors (in blue). We choose the ordered pairs uniformly at random from the set of all ordered pairs of documents by distinct authors. Each panel corresponds to a distinct feature. The means of the distributions of each panel are (a) 0.0828 (black) and 0.240 (blue), (b) 0.167 (black) and 0.433 (blue), (c) 0.149 (black) and 0.275 (blue), and (d) 0.0682 (black) and 0.154 (blue).} 
\label{dist_KL}
\end{figure}

In~\cref{dist_KL}, we show the distribution of KL divergence values between documents by the same authors (in black) and between documents by distinct authors (in blue). For~\cref{consistency_plots}, we use the former to compute author consistency (by taking the mean of the values for each author) and the latter to compute the consistency baseline (by taking the mean of all values). For all features, we see from a Kolmogorov--Smirnov (KS) test that the difference between the empirical distributions is statistically significant. (In all cases, the p-value is less than or equal to $1.218 \times 10^{-79}$.)


\subsection{Author recognition}
\label{sec:author_rec}

We use the classification techniques from~\cref{class_models} to perform author recognition. We show our results using KL divergence (see~\cref{multiclass_NN}) in~\cref{tab:table_neighb} and using neural networks (see~\cref{multiclass_NN2}) in~\cref{tab:table_nn}. In each table, we specify the number of authors (\quoting{No. authors}), the number of documents in the training set (\quoting{Training size}), the number of documents in the testing set (\quoting{Testing size}), the accuracy of the test using various sets of features, and the baseline accuracy (as defined in~\cref{evaluation}). Each row in a table corresponds to an experiment on a set of distinct authors, which we choose uniformly at random. (The set consists of the entire corpus when the number of authors is \nbauthors.) For a given number of authors, we use the same sample across both tables to allow a fair comparison.

We show classification results using KL divergence in~\cref{tab:table_neighb} using each individual frequency feature vector as input. 
As we consider more authors, the accuracy on the testing set tends to decrease significantly. The issue of
developing a method that scales well as one increases the number of authors
is an open problem in author recognition even when using words from
text~\cite{Neal2017}, and we are exploring stylistic signatures from
punctuation only, a much smaller set of information. Remarkably, we are able
to achieve an accuracy of 66\% on a sample of $50$ authors using only the
feature $\p{f}^3$. This is consistent with the plots in~\cref{consistency_plots}, where $\p{f}^3$ gave the best improvement from
the baseline.

We show classification results using a one-layer neural network with $2000$ neurons in~\cref{tab:table_nn} using various sets of input vectors
(which, contrary to when one uses KL divergence, need not be feature vectors that induce probability distributions). We also observe in~\cref{tab:table_nn} that accuracy on the testing set tends to decrease significantly as one increases the number of authors.  Overall, however, the neural network outperforms our KL divergence-based classification. We achieve an accuracy of 62\% when using only $\p{f}^3$
and an accuracy of 72\% when using all feature vectors on a sample of
\nbauthors\ authors (i.e., on the entire corpus). Interestingly, in some of our experiments, using the
features \{$\p{f}^1$, $\p{f}^3$, $\p{f}^4$, $\p{f}^5$\} gives slightly better accuracy
than using all features.

Based on preliminary experiments, our accuracy results in~\cref{tab:table_neighb} and~\cref{tab:table_nn} seem to be robust to (1) different author samples of the same size and (2) different training and testing samples for a given author sample. However, the heterogeneity in accuracy across different author samples of the same size is more pronounced than the heterogeneity that we observe from different training and testing samples for a given author sample, as different author samples can sometimes yield significantly different training and testing set sizes (see~\cref{distribution_books}). Such heterogeneity across different author samples decreases as one increases the number of authors. 

To the best of our knowledge, most attempts thus far at author recognition of literary documents have used data sets that are of significantly smaller scale than our corpus~\cite{Gerlach2018, Neal2017}. One recent example of author analysis from a corpus extracted from Project Gutenberg is the one in Qian \emph{et al.}~\cite{Qian2017}. Their corpus consists of $50$ authors (with their choices of authors based on a popularity criterion) and $900$ single-paragraph excerpts for each author. (For a given author, they extracted their excerpts from several books.) Using word-based features and machine-learning classifiers, they achieved an accuracy of 89.2\% using 90\% of their data for training and 10\% of it for testing.

\begin{table}
\caption{Results of our author-recognition experiments using a classification based on KL divergence (see \cref{multiclass_NN}) for author samples of various sizes and using the individual features $\p{f}^1$, $\p{f}^3$, $\p{f}^4$, and $\p{f}^5$ as input. We measure accuracy as the ratio of correctly assigned documents relative to the total number of documents in the testing set. (See~\cref{evaluation} for a description of the baseline.)
}
\label{tab:table_neighb}
\begin{tabular}{cccccccl}
\hline\\ [-5.3ex]\hline
No. authors & Training size & Testing size & \multicolumn{5}{c}{Accuracy on the testing set} \\ \hline
           &               &   &           $\p{f}^1$       & $\p{f}^3$       & $\p{f}^4$      & $\p{f}^5$    & baseline    \\
10         & 216           & 55           & 0.69       & 0.74       & 0.52      & 0.63  & 0.21   \\
50         & 834           & 209          & 0.54       & 0.66      & 0.30      & 0.31   & 0.029 \\
100        & 2006          & 502          & 0.37       & 0.49      & 0.25      & 0.23   &  0.019   \\
200        & 3549          & 888          & 0.30       & 0.47     & 0.16      & 0.20  & 0.0079  \\
400        &      7439     & 1860          & 0.27       & 0.41      & 0.15      & 0.16  & 0.0047 \\

\hline\\ [-5.3ex]\hline
\end{tabular}
\end{table}

\begin{table}
\caption{Results of our author-recognition experiments using a one-layer, 2000-neuron neural network (see~\cref{multiclass_NN2}) for author samples of various sizes and using different features or sets of features as input: $\p{f}^1$, $\p{f}^3$, $\p{f}^4$, $\p{f}^5$, \{$\p{f}^1$, $\p{f}^3$, $\p{f}^4$, $\p{f}^5$\}, and \{$\p{f}^1$, $\p{f}^2$, $\p{f}^3$, $\p{f}^4$, $\p{f}^5$, $\p{f}^6$\} (which we label as ``all''). We measure accuracy as the ratio of correctly assigned documents relative to the total number of documents in the testing set. (See~\cref{evaluation} for a description of the baseline.)
} \label{tab:table_nn}
\begin{tabular}{cccccccccl}
\hline\\ [-5.3ex]\hline
No. authors & Training size & Testing size & \multicolumn{7}{c}{Accuracy on testing set} \\
\hline
           &               &              & $\p{f}^1$       & $\p{f}^3$       & $\p{f}^4$      & $\p{f}^5$    & \{$\p{f}^1$, $\p{f}^3$, $\p{f}^4$, $\p{f}^5$\} & all  & baseline\\
    
10         & 216           & 55           & 0.89      & 0.93      & 0.64     & 0.80    & 0.89                 & 0.87   & 0.21\\
50         & 834           & 209          &  0.65      & 0.81      &     0.44     & 0.49  & 0.81                 &  0.82   & 0.029  \\
100        & 2006          & 502          & 0.55      & 0.79      & 0.37    & 0.39   & 0.79                  & 0.80   & 0.019\\
200        & 3549          & 888          & 0.46      & 0.71      &    0.23     & 0.32   & 0.71     & 0.75   & 0.0079 \\
400        & 7439          & 1860         & 0.39      & 0.70     &  0.23     & 0.27  &  0.71                  & 0.73  & 0.0047 \\
600        & 11102         & 2776         & 0.37      & 0.70      &  0.21     & 0.25   & 0.61                   & 0.74 & 0.0029\\
651        & 11957         & 2990         &  0.36      &  0.62      & 0.20     &  0.23    &  0.67       & 0.72 & 0.0024\\
\hline\\ [-5.3ex]\hline
\end{tabular}
\end{table}


\section{Case study: Genre analysis}
\label{sec:genres}

\noindent \texttt{"Cut out all those exclamation marks. An exclamation mark is like laughing 
at your own jokes."\newline
\indent \indent \indent \indent \indent \indent \indent \indent \indent \indent \indent \indent
{\normalfont\emdash{}\;Attributed to F. Scott Fitzgerald, as conveyed by Sheilah Graham and Gerold Frank in \textit{Beloved Infidel: The Education of a Woman}, 1958}}
\vspace{2em}

\noindent \texttt{"`Multiple exclamation marks,' he went on, shaking his head, `are a sure sign of a diseased mind.'"\newline
 \indent \indent \indent \indent \indent \indent \indent \indent \indent \indent \indent \indent
{\normalfont\emdash{}\; Terry Pratchett, \textit{Eric}, 1990}}

\medskip
\medskip
\medskip

We now use genres as our classes. Among the $121$ genre (``bookshelf'')
labels that are available in Gutenberg\footnote{Every document in our corpus has at most one genre, but most
documents are not assigned a genre.}, we keep those that include at least $10$ documents. Among the remaining genres, we select $32$ relatively unspecialized genre labels.  We show this final list of genres in~\cref{app:authorsgenres}. This yields a data set with $2413$ documents.


\subsection{Consistency}

\begin{figure}[t]
\centering
\subfigure[feature $\p{f}^1$]{
\includegraphics[width = 3cm]{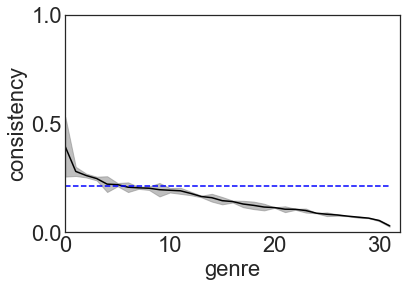}}
\subfigure[feature $\p{f}^3$]{
\includegraphics[width = 3cm]{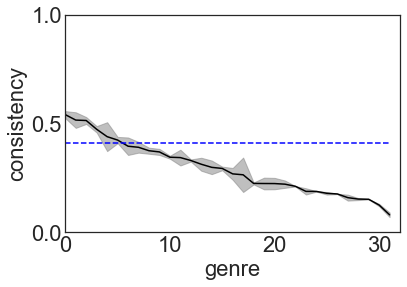}}
\subfigure[feature $\p{f}^4$]{
\includegraphics[width = 3cm]{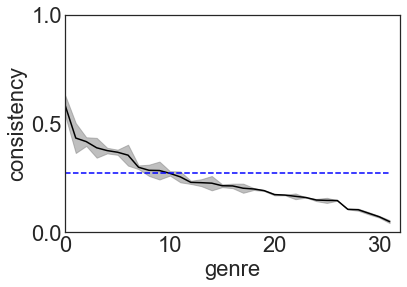}}
\subfigure[feature $\p{f}^5$]{
\includegraphics[width = 3cm]{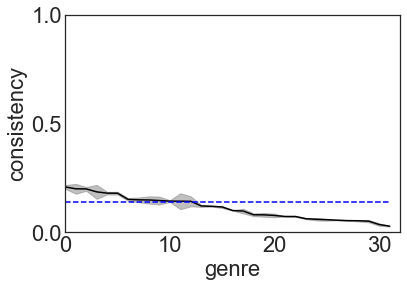}}
\centering\includegraphics[width = 5cm]{figs2/legend_consistency_plot.png}
\caption[]{Evaluation of genre consistency. In each panel, we show the genre consistency (specifically, we use equation~\eqref{consistency}, but with genres, instead of authors) for (a) $\p{f}^1$,
  (b) $\p{f}^3$, (c) $\p{f}^4$, and (d) $\p{f}^5$ as a solid black curve. In gray, we show confidence intervals
  of KL divergence across pairs of documents for
  each genre. To compute the confidence intervals, we assume that the KL divergence across pairs of distinct documents for each genre are normally distributed. 
  There are at least $10$ documents for each genre in our corpus (see the introduction of~\cref{sec:genres}), so the number of KL values across pairs of distinct documents for each genre is at least 90. The dotted blue line indicates a consistency baseline, which we obtain by choosing, uniformly at random, $1000$ ordered pairs of documents from distinct genres and computing the mean KL divergence between the features of these
  document pairs. 
    }
\label{consistency_plots_genre}
\end{figure}

\begin{figure}[t!]
\centering
\subfigure[feature $\p{f}^1$]{
\includegraphics[width = 3cm]{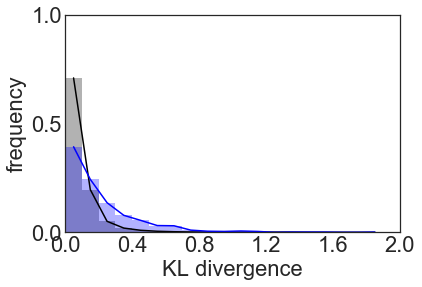}}
\subfigure[feature $\p{f}^3$]{
\includegraphics[width = 3cm]{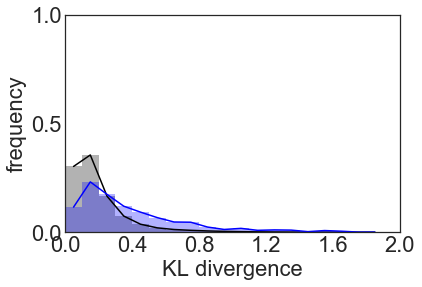}}
\subfigure[feature $\p{f}^4$]{
\includegraphics[width = 3cm]{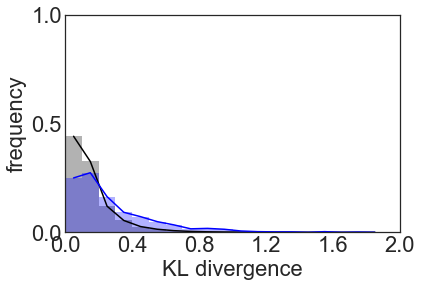}}
\subfigure[feature $\p{f}^5$]{
\includegraphics[width = 3cm]{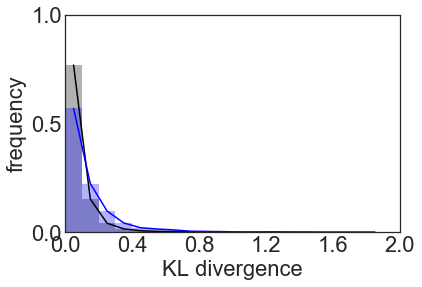}}

\caption[]{Distributions of KL divergence for genre. 
In each panel, we show the distributions of KL divergence between all pairs of documents in the corpus from the same genre (in black) and between $1000$ ordered pairs of documents from distinct genres (in blue). We choose the ordered pairs uniformly at random from the set of all ordered pairs of documents from distinct genres. Each panel corresponds to a distinct feature. The means of the distributions of each panel are (a) 0.102 (black) and 0.215 (blue), (b) 0.206 (black) and 0.412 (blue), (c)  0.154 (black) and 0.272 (blue), and (d) 0.0821 (black) and 0.138 (blue). 
}
\label{dist_KL_genres}
\end{figure}

In~\cref{consistency_plots_genre}, we show consistency plots (of the same type as in~\cref{consistency_plots}), but now we use genres (instead of authors) as our classes. We observe that the KL-divergence consistency relative to the baseline is less pronounced for genres than it was for authors. Nevertheless, most genres are more consistent than the baseline, and the frequency feature vector $\p{f}^{3}$ appears to be the most helpful of our features for evaluating a genre's punctuation style.

In~\cref{dist_KL_genres}, we show the distributions of KL divergence between documents from the same genre (in black) and between documents from different genres (in blue). One can use the former to compute genre consistency in~\cref{consistency_plots_genre} (by taking the mean of the values for each genre) and the latter to compute the consistency baseline in~\cref{consistency_plots_genre} (by taking the mean of all values). For all features, we see from a KS test that the difference between the empirical distributions is statistically significant. (In all cases, the p-value is less than or equal to $2.247 \times 10^{-36}$.)


\subsection{Genre recognition}
\label{sec:genres_rec}

We perform genre recognition using neural networks and show our results in~\cref{tab:table_nn_genre}. We are less successful at genre detection than we were at author detection. This is consistent with our genre consistency plots (see~\cref{consistency_plots_genre}), which indicated a smaller differentiation from the baseline than in our author consistency plots (see~\cref{consistency_plots}). Our highest accuracy for genre recognition is 65\%; we achieve it when using only the feature $\p{f}^{3}$ as input. These observations are robust to different samples of the training and testing sets.

\begin{table}
\caption{Results of our genre-recognition experiments using a one-layer, 2000-neuron neural network (see~\cref{multiclass_NN2}) using different features or sets of features as input: $\p{f}^1$, $\p{f}^3$, $\p{f}^4$, $\p{f}^5$, \{$\p{f}^1$, $\p{f}^3$, $\p{f}^4$, $\p{f}^5$\}, and \{$\p{f}^1$, $\p{f}^2$, $\p{f}^3$, $\p{f}^4$, $\p{f}^5$, $\p{f}^6$\} (which we label as ``all''). We measure accuracy as the ratio of correctly assigned documents relative to the total number of documents in the testing set. (See~\cref{evaluation} for a description of the baseline.)
} \label{tab:table_nn_genre}
\begin{tabular}{cccccccccc}
\hline\\ [-5.3ex]\hline
No. genres & Training size & Testing size & \multicolumn{7}{c}{Accuracy on testing set} \\
\hline
           &               &              & $\p{f}^1$       & $\p{f}^3$       & $\p{f}^4$      & $\p{f}^5$    & \{$\p{f}^1$, $\p{f}^3$, $\p{f}^4$, $\p{f}^5$\} & all  & baseline\\
    
32         & 1930           & 483           & 0.56      & 0.65      & 0.37     & 0.40     & 0.61   & 0.64 & 0.094\\
\hline\\ [-5.3ex]\hline
\end{tabular}
\end{table}


\section{Case study: Temporal analysis}
\label{sec:time}

\texttt{"Whatever it is that you know, or that you don't know, tell me about it. We 
can exchange tirades. The comma is my favorite piece of punctuation and 
I've got all night."\newline
\indent \indent \indent \indent \indent \indent \indent \indent \indent \indent \indent \indent
{\normalfont\emdash{}\;Rasmenia Massoud, \textit{Human Detritus}, 2011}}

\medskip
\medskip
\medskip

\noindent \texttt{"Who gives a @!\#?@! about an Oxford comma? \newline
I've seen those English dramas too \newline
They're cruel" \newline
\indent \indent \indent \indent \indent \indent \indent \indent \indent \indent \indent \indent
{\normalfont\emdash{}\;
Vampire Weekend, \textit{Oxford Comma}, 2008}}

\medskip
\medskip

\begin{figure}[t]
\centering
\subfigure[Birth year and death year]{
\includegraphics[width = 5cm]{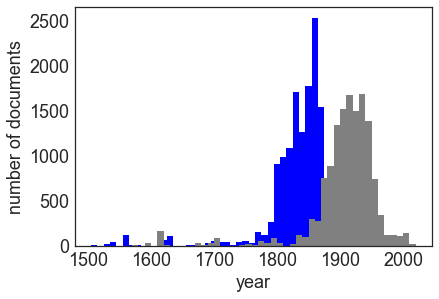}}
\subfigure[Birth year, death year, and middle year]{\includegraphics[width = 5cm]{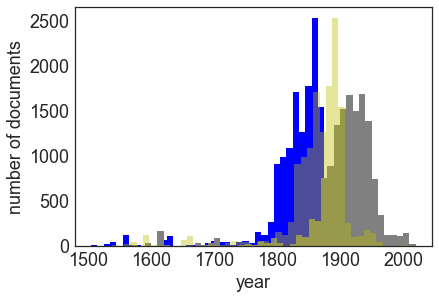}}

\centering\includegraphics[width=5cm]{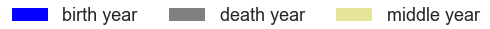}

\caption[]{Distribution of author dates over time in our corpus. The bars represent the number of documents by author birth year (blue) and death year (gray) split into bins, where each bin represents a 10-year period. (We start at 1500.) For ease of visualization, we only show documents for authors who were born in 1500 or later. (Only six of our authors for whom we have birth years were born before 1500.) We determine the ``middle year'' of an author by taking the mean of the birth year and the death year if they are both available. If we know only the birth year, we assume that the middle year of an author is 30 years after the birth year; if we know only the death year, we assume that the middle year is 30 years prior to the death year.
} 
\label{temporal_repartition}
\end{figure}

\begin{figure}[t]
\centering
\subfigure[Punctuation marks over time]{
\includegraphics[width = 7cm]{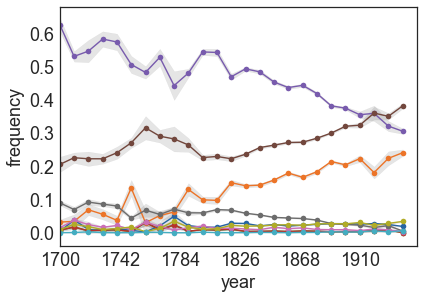}}

\subfigure[Quotation mark, period, and comma]{
\includegraphics[width = 4cm]{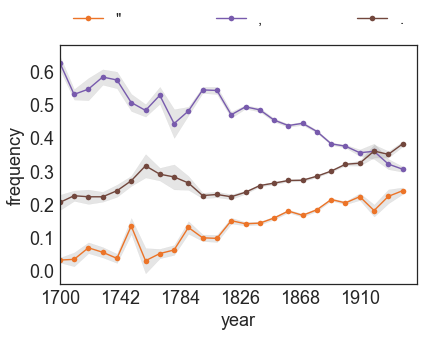}}
\hspace{1mm}
\subfigure[Exclamation mark and semicolon]{
\includegraphics[width = 4cm]{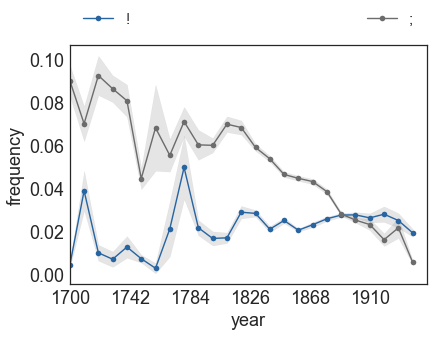}}
\hspace{1mm}
\subfigure[Left parenthesis, right parenthesis, colon, question mark, and ellipsis]{
\includegraphics[width = 4cm]{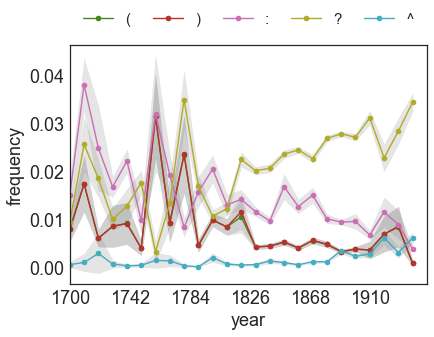}}

\centering\includegraphics[width = 7cm]{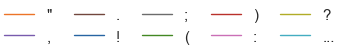}

\caption[]{Mean frequency of punctuation marks versus the middle years of authors. Recall that $\p{f}^{1,k}$
is the frequency of punctuation marks for document $k$. We bin middle years into 10-year periods that start at 1700. In (a), we show the temporal evolution of all punctuation marks. For clarity, we also separately plot (b) the three punctuation marks with the largest frequencies in the final year of our data set, (c) the next two most-frequent punctuation marks, and (d) the remaining punctuation marks. The gray shaded area indicates confidence intervals. To compute the confidence intervals, we assume that the values of $\p{f}^{1,k}$ are normally distributed for each year. 
} 
\label{freq_pun_overtime}
\end{figure}

\begin{figure}[t]
\centering
\subfigure
{
\includegraphics[width = 7cm]{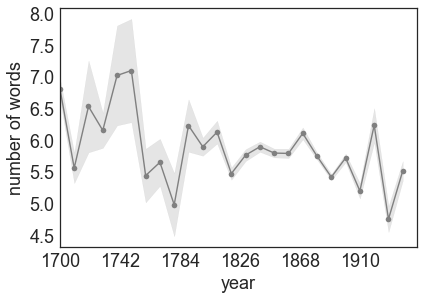}}

\centering\includegraphics[width = 7cm]{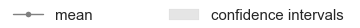}

\caption[]{Temporal evolution of the mean number of words between two consecutive punctuation marks (i.e., $\mathbb{E}\left[\p{f}^{5,k}\right]$ from formula \eqref{punctuation_rate}) versus author middle years, which we bin into 10-year periods that start at 1700. The gray shaded area indicates confidence intervals. To compute the confidence intervals, we assume that the values of $\mathbb{E}\left[\p{f}^{5,k}\right]$ are normally distributed for each year. This reflects how the punctuation rate in our corpus has changed over time.
} 
\label{7_overtime_3plots}
\end{figure}

\begin{figure}[t]
\centering
\subfigure[H. G. Wells over time]{
\includegraphics[width = 4cm]{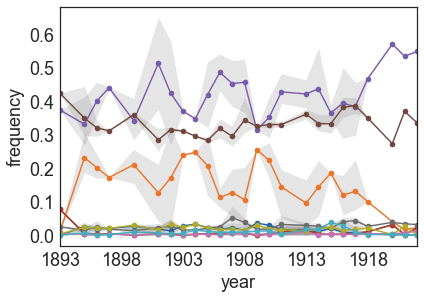}}
\subfigure[A. M. Fleming over time]{
\includegraphics[width = 4cm]{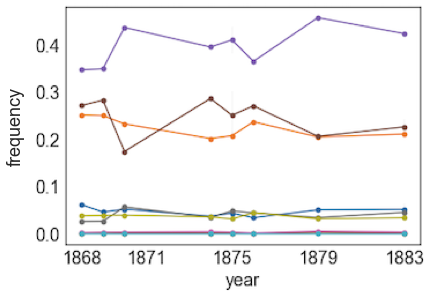}}
\subfigure[C. Dickens over time]{
\includegraphics[width = 4cm]{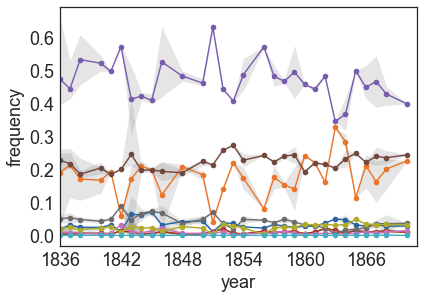}}
\centering\includegraphics[width = 7cm]{figs2/freq1_overtime_legend123.png}
\caption[]{Mean frequency of punctuation marks versus publication
  date for works by (a) Herbert George Wells, (b) Agnes May Fleming, and (c) Charles Dickens. Recall that $\p{f}^{1,k}$
is the frequency of punctuation marks for document $k$. The gray shaded area indicates the minimum and maximum value of $\p{f}^{1,k}$ for each year. (Because of the small sample sizes, we do not show confidence intervals.) 
}
\label{shak_wells_overtime}
\end{figure}

We perform experiments to obtain preliminary insight into how punctuation has changed over time. In our corpus, we have access to the birth year and death year of $614$ and $615$ authors, respectively, of the \nbauthors\ total authors. We have both the birth and death years for $607$ authors. In~\cref{temporal_repartition}, we show the distribution of the number of documents by author birth year, death year, and ``middle year''.\footnote{We use ``middle year'' as a proxy for ``publication year'', which is unavailable in the metadata of Project Gutenberg. Our results are qualitatively similar when we use birth year or death year instead of middle year.} (See the caption of~\cref{temporal_repartition} for the definition of middle year.) We restrict our analysis to authors with a middle year between 1500 and 2012. Of the authors for whom we possess either a birth year or a death year, $616$ of them have a middle year between 1500 and 2012. We show the evolution of punctuation marks over time for these $616$ authors in~\cref{freq_pun_overtime} and~\cref{7_overtime_3plots}, and we examine the punctuation usage of specific authors over time in~\cref{shak_wells_overtime}. Based on our experiments, it appears from~\cref{freq_pun_overtime} that the use of quotation marks and periods has increased over time (at least in our corpus), but that the use of commas has decreased over time. Less noticeably, the use of semicolons has also decreased over time.\footnote{See~\cite{Watson2019} for a ``biography'' of the semicolon, which reportedly was invented in 1494.} In~\cref{7_overtime_3plots}, we observe that the punctuation rate (given by the formula~\eqref{punctuation_rate}) tends to decrease over time in our corpus. However, this observation requires further statistical testing, especially given the large variance in ~\cref{7_overtime_3plots}. Because of our relatively small number of documents per author and the uneven distribution of documents in time, our experiments in~\cref{shak_wells_overtime} give only preliminary insights into the temporal evolution of punctuation, which merits a thorough analysis with a much larger (and more appropriately sampled) corpus. Nevertheless, this case study illustrates the potential for studying the temporal evolution of punctuation styles of authors, genres, and literature (and other text) more generally.

\section{Conclusions and Discussion}
\label{sec:conc}

\noindent \texttt{"La punteggiatura \`{e} come l'elettroencefalogramma di un cervello che sogna 
\emdash{} non d\`{a} le immagini ma rivela il ritmo del flusso sottostante." \newline
\indent \indent \indent \indent \indent \indent \indent \indent \indent \indent \indent \indent {\normalfont\emdash{}\;Andrea Moro, \textit{Il Segreto di Pietramala}, 2018}}
\vspace{2em}

We have explored whether punctuation is a sufficiently rich stylistic feature to distinguish between different authors and between different genres, and we have also examined how it has evolved over time. Using a large corpus of documents from Project Gutenberg, we observed that simple punctuation-based quantitative features (which account for both frequency and order) can distinguish accurately between the styles of different authors. These features can also help distinguish between genres, although less successfully than for authors. One feature, which we denote by $\p{f}^3$, measures the frequency of successive punctuation marks (and thereby accounts for the order in which punctuation marks appear). Among the features that we studied, it revealed the most information about punctuation style across all of our experiments. It is worth noting that, unlike $\p{f}^2$, which also accounts for the order of punctuation marks, $\p{f}^3$ gives less weight to rare events and more weight to frequent events (see~\cref{NTM}). This characteristic of $\p{f}^3$, coupled with the fact that it accounts for the order of punctuation marks, may explain some of its success in our experiments. It would be interesting to investigate whether particular entries of $\p{f}^3$ have more predictive power than others, and it is also worth exploring accuracy as a function of the length of the punctuation sequences that one extracts from a document. The latter may shed light on how much of a ``punctuation signal'' is necessary to determine an author's stylistic footprint. In preliminary explorations, we also observed changes in punctuation style across time, but it is necessary to conduct more thorough investigations of temporal usage patterns. 

To assess whether our observations extend beyond our Project Gutenberg corpus, it is necessary to conduct further experiments (e.g., on a larger corpus, across different e-book sources, and so on). For example, it is desirable to repeat our analysis using the \quoting{Text data} level of granularity in the recently introduced Standardized Project Gutenberg Corpus~\cite{Gerlach2018}. We also reiterate that although we associate documents to authors throughout our paper as an expository shortcut, authors and editors both influence a document's writing and punctuation style, and we do not distinguish between the two in our analysis. It would be interesting (although daunting and computationally challenging for Project Gutenberg) to try to gauge whether and how much different editors affect authorial style.\footnote{Such an analysis may be easier with academic papers, as one can compare papers on arXiv to their published versions.} 
It is also worth reiterating that Project Gutenberg has limitations with the cleanliness of its data. (See our discussion in~\cref{sec:database1} for examples of such issues.) These issues may be inherited from the e-books themselves, they may be related to how the documents were entered into Project Gutenberg, or both issues may be present. Although we extensively clean the Project Gutenberg data to ameliorate some of its limitations, important future work is comparing documents that one extracts from Project Gutenberg with the same documents from other data sources.

Our framework allows the exploration of numerous other fascinating ideas. For example, we expect it to be fruitful to examine higher-order categorical Markov chains when accounting for punctuation order. Additionally, we look forward to extensions of our work that explore other features, such as the number of words between elements in ordered pairs of punctuation marks (even when they are not successive) and different ways of measuring punctuation frequency~\cite{Grieve2007} and sentence length~\cite{Vieira2018}), and that try to quantify how large a sample of a document is necessary to correctly identify its features of punctuation style. If this size is sufficiently small, it may even be possible to identify punctuation style from collections of short text (such as tweets from politicians with limited coherence). It is also likely to be useful to exploit more sophisticated machine-learning classifiers that can take raw punctuation sequences (rather than features that one produces from them) as input and exploit \quoting{long-range correlations}~\cite{Ebeling1993} between punctuation marks.

Building on our analysis, it will be interesting to investigate other aspects of stylometry --- such as author pacing or the influence on an author of gender, culture, other demographics, local history, or other aspects of humanity --- and to compare the results of punctuation-based stylometry with existing (word-based) approaches in NLP on the same tasks. One can also explore how successful punctuation-based features are at plagiarism detection and investigate whether the punctuation in a part of a document (e.g., one chapter) is representative of the punctuation in a whole document. Further investigations of a punctuation-based approach to stylometry also provide an opportunity to apply other methods for analyzing categorical time series (e.g., an extension of rough-path signatures~\cite{Lyons2014, Chevyreva2016} to categorical time series).

On a more general front, relevant stylometric applications include analysis of stylistic differences in punctuation between politicians from different political parties \cite{calhoun2} and comparisons between different editions of the same book. It would also be interesting to explore the effects of an editor's or journal's style on documents by a given author (an especially relevant study, in light of the potential to confound such contributions in corpuses like Project Gutenberg), as well as the effects of a translator's style on documents. We envisage that the latter application is particularly well-suited to punctuation-based stylometry, as punctuation marks depend far less than words on the specific choice of language. We also expect there to be commercial applications (e.g., using online data sources) of time-series analysis of symbols without the use of words.


\section*{Acknowledgements}

The original inspiration for this project was Adam Calhoun's blog entry \cite{Calhoun2016blog} and its striking visualizations of punctuation sequences. We thank Mariano Beguerisse D\'iaz, Arthur Benjamin, Bryan Bischof, Chris Brew, Cynthia Gong, Joanna Innes, Jalil Kazemitabar, Aisling Kelliher, Terry Lyons, Ursula Martin, Stephen Pulman, Massimo Stella, Adam Tsakalidis, Dmitri Vainchtein, Bo Wang, and two anonymous referees for helpful comments. Other attendees at SDH's 60th birthday workshop (see \url{https://www.maths.ox.ac.uk/groups/mathematical-finance/sam-howisons-60th-birthday-workshop-2018}) also made helpful comments. For part of this project, MB was supported by The Alan Turing Institute under EPSRC grant EP/N510129/1. MAP and SDH thank their students and postdocs for putting up with many long discussions about punctuation when they perhaps should have been discussing other elements of their scholarship. (It was inevitable that we would eventually write an article like this.) MAP thanks SDH for his collaboration and friendship, and he wishes him a very happy birthday filled with British spelling, the word ``which'' (and occasionally ``that''), and minimal commas (and parenthetical remarks). 


\appendix


\section{Author and genre lists}
\label{app:authorsgenres}

\noindent \texttt{"Mr Speaker, I said the honourable Member was a liar it is true and I am 
sorry for it. The honourable Member may place the punctuation where he 
pleases." \newline
\indent \indent \indent \indent \indent \indent \indent \indent \indent \indent \indent \indent	{\normalfont\emdash{}\;Attributed to Richard Brinsley Sheridan (1751{\normalfont --}1816), responding 
to a rebuke from the Chair for calling a fellow Member of Parliament a 
liar.}}
\vspace{2em}

In \cref{author-table}, we list the authors that we use in our study. We order them based on their $\p{f}^3$ consistency, where smaller numbers indicate greater consistency. (See~\eqref{consistency} for the definition of ``consistency''.) The author order proceeds down the first column and then down the second column. We structure each row as follows: ``Author name (number of documents by that author in our corpus, test-set size for our experiments on the full corpus with the full set of features, author $\p{f}^3$ consistency in our corpus, author accuracy on test set)''. Accuracy values that are closer to $1$ indicate that we correctly assign a larger fraction of books by that author. (See \eqref{evaluation} for the definition of ``accuracy''.) The designation ``NA'' indicates that an author is not in the test set. We number each row in \cref{author-table} to facilitate the referencing of specific authors. One number references two distinct authors (with one in each column), and we increment the row number from page to page in a way that accounts for the number of authors in the second column.

In \cref{genre-table2}, we list the genres that we use in our study. We order them based on their $\p{f}^3$ consistency. The genre order proceeds down the first column and then down the second column. We structure each row as follows: ``Genre (number of documents in the genre, test-set size for our experiments on the full corpus with the full set of features, author $\p{f}^3$ consistency in our corpus, genre accuracy on test set)''. Consistency values that are closer to $0$ correspond to genres that are more consistent, and accuracy values that are closer to $1$ indicate that we correctly assign a larger fraction of books of that author. We number each row in \cref{genre-table2} to facilitate the referencing of specific genres. One number references two distinct genres (with one in each column).

\begin{table}
\scriptsize
\caption{The authors that we use in our study.} 
\label{author-table}
\vspace{5mm}
\begin{tabular}{llll}
   & author (No. documents - test size - consistency - accuracy)      & author (No. documents - test size - consistency - accuracy)        &                                      \\ \hline
0  & Matthews, Stanley R. (32 - 5.0 - 0.018 - 1.0)                    & Werner, E. (18 - 2.0 - 0.053 - 1.0)                                &                                      \\
1  & Hill, Grace Brooks (11 - 2.0 - 0.02 - 0.0)                       & Kyne, Peter B. (Peter Bernard) (10 - 2.0 - 0.054 - 0.0)            &                                      \\
2  & Dell, Ethel M. (Ethel May) (16 - 5.0 - 0.02 - 0.8)               & Wood, Henry, Mrs. (24 - 6.0 - 0.055 - 1.0)                         &                                      \\
3  & Goodwin, Harold L. (Harold Leland) (13 - 2.0 - 0.021 - 1.0)      & King, Charles (27 - 6.0 - 0.055 - 1.0)                             &                                      \\
4  & Young, Clarence (23 - 7.0 - 0.023 - 0.857)                       & Bassett, Sara Ware (16 - 3.0 - 0.055 - 0.333)                      &                                      \\
5  & Hancock, H. Irving (Harrie Irving) (40 - 9.0 - 0.024 - 1.0)      & Abbott, John S. C. (John Stevens Cabot) (23 - 7.0 - 0.056 - 1.0)   &                                      \\
6  & Wirt, Mildred A. (Mildred Augustine) (30 - 5.0 - 0.024 - 1.0)    & Gregory, Jackson (10 - NA - 0.056 - NA)                          &                                      \\
\end{tabular}
\end{table}

\begin{table}
\scriptsize
\vspace{5mm}
\begin{tabular}{lll}
    & author (No. documents - test size - consistency - accuracy)         & author (No. documents - test size - consistency - accuracy)         \\ \hline

150 & Sheckley, Robert (18 - 6.0 - 0.03 - 0.667)                          & Cawein, Madison Julius (19 - 1.0 - 0.039 - 1.0)                     \\
151 & Williamson, C. N. (Charles Norris) (19 - 5.0 - 0.03 - 0.4)          & King, Basil (10 - 4.0 - 0.039 - 1.0)                                \\
152 & Vandercook, Margaret (24 - 4.0 - 0.03 - 1.0)                        & J{\'o}kai, M{\'o}r (28 - 9.0 - 0.039 - 0.444)                               \\
153 & Schubin, Ossip (10 - 2.0 - 0.03 - 0.0)                              & Schmitz, James H. (10 - NA - 0.039 - NA)                          \\
154 & Mangasarian, M. M. (Mangasar Mugurditch) (12 - NA - 0.03 - NA)    & Rohmer, Sax (17 - 1.0 - 0.039 - 0.0)                                \\
155 & Chambers, Robert W. (Robert William) (43 - 9.0 - 0.03 - 0.889)      & Sue, Eug{\`e}ne (44 - 11.0 - 0.04 - 0.818)                              \\
156 & Heyse, Paul (10 - 4.0 - 0.03 - 0.25)                                & Reynolds, Mack (24 - 5.0 - 0.04 - 1.0)                              \\
\end{tabular}
\end{table}

\begin{table}
\vspace{5mm}
\scriptsize
\begin{tabular}{lll}
    & author (No. documents - test size - consistency - accuracy)      & author (No. documents - test size - consistency - accuracy)            \\ \hline
300 & Wiggin, Kate Douglas Smith (33 - 6.0 - 0.052 - 0.667)            & Chesterfield, Philip Dormer Stanhope, Earl of (12 - 2.0 - 0.069 - 1.0) \\
301 & Samachson, Joseph (12 - 1.0 - 0.052 - 0.0)                       & Collins, Wilkie (35 - 5.0 - 0.069 - 0.8)                               \\
302 & Mitton, G. E. (Geraldine Edith) (12 - 2.0 - 0.052 - 0.0)         & Atherton, Gertrude Franklin Horn (25 - 5.0 - 0.07 - 0.8)               \\
303 & Curwood, James Oliver (27 - NA - 0.052 - NA)                   & Huneker, James (11 - 1.0 - 0.07 - 0.0)                                 \\
304 & Kjelgaard, Jim (11 - 4.0 - 0.052 - 0.75)                         & Swift, Jonathan (16 - 1.0 - 0.07 - 0.0)                                \\
305 & Crawford, F. Marion (Francis Marion) (47 - 11.0 - 0.052 - 0.818) & Huxley, Thomas Henry (48 - 13.0 - 0.07 - 0.692)                        \\
306 & Nourse, Alan Edward (23 - 3.0 - 0.053 - 0.667)                   & Rinehart, Mary Roberts (29 - 6.0 - 0.07 - 0.333)                       \\
\end{tabular}
\end{table}

\begin{table}
\vspace{5mm}
\scriptsize
\begin{tabular}{lll}
    & author (No. documents - test size - consistency - accuracy)          & author (No. documents - test size - consistency - accuracy)       \\ \hline
450 & Hegel, Georg Wilhelm Friedrich (10 - 2.0 - 0.192 - 1.0)              & Moli\`ere (20 - 4.0 - 0.244 - 0.75)                \\
451 & Daudet, Alphonse (17 - 3.0 - 0.193 - 0.333)                          & Fletcher, John (15 - 4.0 - 0.244 - 0.0)                           \\
452 & Brinton, Daniel G. (Daniel Garrison) (18 - 3.0 - 0.193 - 0.667)      & Lebert, Marie (15 - 4.0 - 0.247 - 0.75)                           \\
453 & France, Anatole (31 - 3.0 - 0.194 - 0.667)                           & Schoolcraft, Henry Rowe (13 - 3.0 - 0.247 - 0.333)                \\
454 & Hakluyt, Richard (15 - 3.0 - 0.195 - 1.0)                            & Saltus, Edgar (13 - 4.0 - 0.249 - 0.25)                           \\
455 & Duellman, William Edward (12 - 2.0 - 0.195 - 0.5)                    & Ballou, Maturin Murray (19 - 5.0 - 0.249 - 0.4)                   \\
456 & Janifer, Laurence M. (12 - 2.0 - 0.196 - 1.0)                        & Page, Thomas Nelson (24 - 6.0 - 0.25 - 0.5)                       \\
\end{tabular}
\end{table}

\begin{table}
\vspace{5mm}
\scriptsize
\begin{tabular}{lll}
    & author (No. documents - test size - consistency - accuracy)                             &  
    {\color{white} 
    author (No. documents - test size - consistency - accuracy)} \\ \hline
600 & Jackson, Helen Hunt (13 - 2.0 - 0.369 - 0.0)                                            &                                                             \\
601 & Crane, Walter (17 - 4.0 - 0.37 - 0.5)                                                   &                                                             \\
602 & Goethe, Johann Wolfgang von (15 - 2.0 - 0.371 - 0.5)                                    &                                                             \\
603 & Bj\o rnson, Bj\o rnstjerne (16 - 3.0 - 0.372 - 1.0) &                                                             \\
604 & Carroll, Lewis (19 - 4.0 - 0.373 - 1.0)                                                 &                                                             \\
605 & Kipling, Rudyard (44 - 10.0 - 0.373 - 0.7)                                              &                                                             \\
606 & Riley, James Whitcomb (17 - 2.0 - 0.377 - 0.5)                                          &                                                             \\
\end{tabular}
\end{table}

\begin{table}
\scriptsize
\caption{The genres that we use in our study.}
\label{genre-table2}
\vspace{5mm}
\begin{tabular}{lll}
   & genre (No. documents - test size - consistency - accuracy) & genre (No. documents - test size - consistency - accuracy) \\ \hline
0  & World War II (11 - 6 - 0.08 - 0.667)                       & World War I (57 - 17 - 0.294 - 0.235)                      \\
1  & Crime Fiction (27 - 7 - 0.125 - 0.857)                     & Art (14 - 1 - 0.299 - 1.0)                                 \\
2  & Historical Fiction (263 - 48 - 0.152 - 0.833)              & Animal (16 - 2 - 0.313 - 1.0)                              \\
3  & Western (76 - 18 - 0.153 - 0.611)                          & Children's Literature (158 - 26 - 0.33 - 0.462)            \\
4  & Horror (16 - 2 - 0.159 - 0.0)                              & Classical Antiquity (13 - 3 - 0.344 - 0.0)                 \\
5  & Children's Book Series (354 - 75 - 0.176 - 0.853)          & US Civil War (78 - 13 - 0.345 - 0.462)                     \\
6  & Adventure (37 - 9 - 0.179 - 0.333)                         & Christmas (44 - 8 - 0.37 - 0.125)                          \\
\end{tabular}
\end{table}



 
\end{document}